# An automated pipeline for the discovery of conspiracy and conspiracy theory narrative frameworks: Bridgegate, Pizzagate and storytelling on the web

**Timothy R. Tangherlini**[1☯*], **Shadi Shahsavari**[2☯], **Behnam Shahbazi**[2☯], **Ehsan Ebrahimzadeh**[2], **Vwani Roychowdhury**[2☯]

**1** Scandinavian Section, UCLA, Los Angeles, CA, United States of America, **2** Electrical and Computer Engineering, UCLA, Los Angeles, CA, United States of America

☯ These authors contributed equally to this work.
* tango@g.ucla.edu

**Data Availability Statement:** Although the authors cannot make their study's data publicly available at the time of publication, all authors commit to make the data underlying the findings described in this study fully available without restriction to those who request the data, in compliance with the PLOS Data Availability policy. For data sets involving personally identifiable information or other sensitive data, data sharing is contingent on the data being handled appropriately by the data requester and in accordance with all applicable

## Abstract

Although a great deal of attention has been paid to how conspiracy theories circulate on social media, and the deleterious effect that they, and their factual counterpart conspiracies, have on political institutions, there has been little computational work done on describing their narrative structures. Predicating our work on narrative theory, we present an automated pipeline for the discovery and description of the generative narrative frameworks of conspiracy theories that circulate on social media, and actual conspiracies reported in the news media. We base this work on two separate comprehensive repositories of blog posts and news articles describing the well-known conspiracy theory Pizzagate from 2016, and the New Jersey political conspiracy Bridgegate from 2013. Inspired by the qualitative narrative theory of Greimas, we formulate a graphical generative machine learning model where nodes represent actors/actants, and multi-edges and self-loops among nodes capture context-specific relationships. Posts and news items are viewed as samples of subgraphs of the hidden narrative framework network. The problem of reconstructing the underlying narrative structure is then posed as a latent model estimation problem. To derive the narrative frameworks in our target corpora, we automatically extract and aggregate the actants (people, places, objects) and their relationships from the posts and articles. We capture context specific actants and interactant relationships by developing a system of supernodes and subnodes. We use these to construct an actant-relationship network, which constitutes the underlying generative narrative framework for each of the corpora. We show how the Pizzagate framework relies on the conspiracy theorists' interpretation of "hidden knowledge" to link otherwise unlinked domains of human interaction, and hypothesize that this multi-domain focus is an important feature of conspiracy theories. We contrast this to the single domain focus of an actual conspiracy. While Pizzagate relies on the alignment of multiple domains, Bridgegate remains firmly rooted in the single domain of New Jersey politics. We hypothesize that the narrative framework of a conspiracy theory might stabilize quickly in contrast to the narrative framework of an actual conspiracy, which might develop more local requirements. Data are held at the UCLA dataverse repository (doi:10.25346/S6/VTFXSX). Requests for data access can be sent to Timothy R. Tangherlini (tango@g.ucla.edu), Vwani Roychowdhury (vwani@g.ucla.edu), or the dataverse curation team at the UCLA library (datascience@library.ucla.edu). In the application, specify your affiliation and the research purpose of access to the materials; include a copy of your IRB approval or exemption if applicable.

**Funding:** The author(s) received no specific funding for this work.

**Competing interests:** The authors have declared that no competing interests exist.

slowly as revelations come to light. By highlighting the structural differences between the two narrative frameworks, our approach could be used by private and public analysts to help distinguish between conspiracy theories and conspiracies.

## Introduction

Conspiracy theories and their factual counterpart, conspiracies, have long been studied by scholars from a broad range of disciplines, including political science [1][2][3][4], philosophy [5], psychology [6][7][8][9][10][11][12][13][14][15], law [16], sociology [17][18], linguistics and language use [19][20], folklore [21][22] and history [23][24][25]. The recent amplification of conspiracy theories on social media and internet forums has led to an increase in attention paid to how these stories circulate [26][27][28][29][30], and engendered discussions of the impact these stories may have on decision making [31][32][33]. Rosenblum and Muirhead suggest that the corrosive nature of conspiracism intrinsic to these stories, their impact on Democracy writ large and, more narrowly, on democratic institutions such as a free, independent press, warrant significant study [1].

Despite the attention that conspiracy theories have drawn, little attention has been paid to their narrative structure, although numerous studies recognize that conspiracy theories rest on a strong narrative foundation [14][16][21] or that there may be methods useful for classifying them according to certain narrative features such as topics or motifs [19][34].

Part of the challenge of studying the narrative structure of conspiracy theories stems from the fragmentary manner in which they are often discussed. Although the rise of social media has provided a convenient arena for studying the emergence of these narratives, the at times fleeting nature of communications on the forums or platforms where conspiracy theories grow and circulate makes conspiracy theories difficult to identify, track and study [30]. Posts to forums where conspiracy theories take shape can be incomplete or allude to offline conversations or inaccessible websites [19][30][35]. In addition, conspiracy theorists frequently refer to events, places, and people with coded or otherwise hard to decipher language. Occasionally, entire discussion forums disappear, either because they are abandoned, deleted or otherwise overwritten [36][37][38]. Consequently, determining the underlying narrative framework of a conspiracy theory—its cast of characters, the relationships between those characters, the contexts in which those relationships arise, and the previously hidden events the interpretation of which comprise the conspiracy theory's action—is difficult. Yet understanding the underlying narrative framework which, in the case of conspiracy theories, is often the work of multiple people negotiating the boundaries of the narrative through repeated, albeit brief, interactions, can provide significant insight into the various sources of threat imagined by the conspiracy theorists, identify allegedly hidden or special knowledge on which their theorizing rests, and detail the strategies they suggest to counteract the threats encoded in the conspiracy theory [14][39][40]. These strategies can have real world consequences, as evidenced by the case of Edgar Welch, who opened fire with a rifle in a Washington DC area family restaurant while "investigating" the claims of the Pizzagate conspiracy theory [41]. Therefore, the structural understanding of both conspiracy theories and conspiracies provided by our pipeline can be of significant value to many groups, including those charged with ensuring public safety.

In the following work, we differentiate between conspiracy theories, which are largely fictional accounts that comprise, "scenarios made up of many beliefs and narratives which are accepted on faith and used to link and give meaning to stressful events" [21], and actual conspiracies, which are factual events comprised of malign actors working covertly, often in an extralegal manner, to effect some sort of outcome beneficial to those actors [2]. While conspiracies actually exist and are sometimes uncovered, conspiracy theories do not need to have any basis in truth. They are by their very nature always uncovered, since they only exist in narrative. One of the motivating questions for our work is whether the narrative frameworks of conspiracy theories differ in any consistent and identifiable manner from those of actual conspiracies. We delineate two main challenges: First, given the benefits of network representations of narratives identified by narrative scholars [42][43], can we devise automated methods to discover the underlying narrative framework in discussions about a conspiracy theory or reporting about a conspiracy, and represent that as a network graph? Second, can we determine any structural differences in their narrative frameworks?

To meet these challenges, we developed a pipeline of interlocking computational methods to determine the generative narrative framework undergirding a knowledge domain or connecting several knowledge domains. We base the concept of knowledge domain on George Boole's notion of discourse, and his key observation that, "In every discourse, whether of the mind conversing with its own thoughts, or of the individual in his intercourse with others, there is an assumed or expressed limit within which the subjects of its operation are confined" [44]. Extending earlier work on anti-vaccination blog posts and the legend/rumor genre in general [35][39][45][46], we refine an actant-relationship model inspired by Algirdas Greimas's actantial model [47][48]. For Greimas, the model consists of three main components: actants (people, places, things), relationships between actants, and a sequencing of these relationships [49][50]. Operationalizing this approach allows us to determine an actant-relationship graph which describes the generative narrative framework for a particular domain [35][39], and is in keeping with narrative theory that proposes that, in any narrative domain, there are limits on the admissible actants and the relationships between them [51]. This approach also aligns well with work on narrative structure that explains the benefits of representing narratives as graphs: "By representing complex event sequences as networks, we are easily able to observe and measure structural features of narratives that may otherwise be difficult to see" [42]. Since our approach is computational, it allows for a finer grained representation of the numerous actants and their myriad relationships than hand-drawn graphs. Consequently, our work supports a "macroscopic" approach to narrative analysis that can, "provide a 'vision of the whole,' helping us 'synthesize' the related elements and detect patterns, trends, and outliers while granting access to myriad details" [52].

In the context of conspiracy theories, narrative network graphs have been popularized by the artist Mark Lombardi and, more recently, Dylan Louis Monroe, whose Q-Anon "map" has achieved considerable distribution on the internet and through art exhibitions, including one at the Metropolitan Museum of Art [53][54]. These types of graphs have also received broad scale exposure through the hand-drawn illustrations featured in the New York Times, for example those detailing the Pizzagate conspiracy theory and the Bridgegate conspiracy, which were published to aid readers trying to make sense of these complex narratives [55][56]. These illustrations also provide us with two clear target narratives for analysis, one a conspiracy theory and one a conspiracy; the illustrations can serve as expert labeled graphs against which to validate our results.

Any storytelling event, such as a blog post or a news report, activates a subgraph comprising a selection of actants (nodes) and relationships (edges) from the narrative framework. The more often an actant-relationship is activated, the more likely it is to be activated in future tellings, with additions and deletions becoming less and less common, a phenomenon described by Anderson's law of self-correction [57]. As more people contribute stories or parts of stories, the narrative framework is likely to stabilize since the nodes and edges become more heavily weighted each time they are activated. Even though the story may never be told in full, the members of the community circulating these stories and story fragments collectively recognize the immanent narrative that provides a framework for understanding the current story and the creation of additional stories [58]. This concept of the immanent narrative supports existing work on conspiracy theories, with Samory and Mitra pointing out that, "conspiracy theories are often collages of many smaller scale theories" [19].

Recognizing that people rarely tell complete stories [59], and that random sampling from an internet forum would potentially miss important actants and their relationships, we present an automated pipeline for aggregating actants and relationships from as comprehensive a collection as possible of posts or articles about a particular conspiracy or conspiracy theory in order to discover the underlying narrative framework. The actants are combined into supernodes consisting of subnodes that represent the context dependent relationships of that actant. The relationships constitute the edges that connect nodes, which are in turn ranked based on their significance. The resulting network graph comprises the generative narrative framework. The pipeline we have developed is domain independent, and enables the automatic discovery of the underlying narrative framework or frameworks for any domain or group of domains. As such, the pipeline is agnostic to input, and does not presuppose a single narrative framework for any target corpus, nor does it presuppose any classification into narrative classes (e.g. conspiracy theory or conspiracy) for that corpus.

In this work, we use the pipeline to discover the narrative framework for the Pizzagate conspiracy theory, and contrast it with the narrative framework of the Bridgegate conspiracy. We hypothesize that the Pizzagate framework, despite relying on multiple domains of knowledge, reaches a stable state relatively quickly and then becomes resistant to additions or deletions, except in certain circumstances when it expands quickly by aligning nodes and relationships from additional domains to those already contributing to the conspiracy theory. This rapid growth caused by the alignment of additional domains reflects the phenomenon of a monological belief system, which various political psychologists have noted is an important component of conspiratorial thinking and often exhibited by adherents of conspiracy theories [6][15][60]. These alignments frequently occur, according to the conspiracy theorists, through their interpretation of hidden knowledge accessible, at least initially, only to them. These interpretations manifest as heretofore unknown relationships (edges) between actants that cross domains, or the identification of an actant (node or supernode) from one domain in a separate domain where they were not otherwise known to be active. For example, in the Pizzagate conspiracy theory, based on an inventive reading of John Podesta's emails hacked from the DNC servers, Hillary Clinton is discovered by the conspiracy theorists to be an actant not only in the domain of politics, but also in the domain of human trafficking.

By way of contrast, Bridgegate, while broad in its scope with a large number of actants and interactant relationships, is confined, as most actual conspiracies may be, to a single domain of human interaction, in this case New Jersey politics and, in particular, the fraught relationship between state and local governments. Despite the limited single domain purview of the conspiracy, the narrative framework was still in flux nearly seven years after the initial conspiracy was uncovered, and the number of actants and relationships discovered were far greater than those discovered for Pizzagate.

## Data

Data for this study were derived from two online repositories archived by the UCLA library. We received an exemption from UCLA's Institutional Review Board (UCLA IRB Exemption 19-001257) to make use of this data, as neither we nor the UCLA library had access to any personal identifying information (PII) nor any key to access such information. To ensure that we were in compliance with IRB approvals, prior to beginning our work, we confirmed that the datasets we accessed from the library contained no PII.

For the Pizzagate conspiracy theory, the library based their collection on the Reddit subreddit, r/pizzagate. As with many other conspiracy theories, the community discussing and negotiating the boundaries of Pizzagate archived their own discussions, particularly given their legitimate concern that Reddit was considering banning their subreddit [36]. The Pizzagate community moved their discussions to Voat in the aftermath of Reddit's decision, and continued their discussions on v/pizzagate. This data collection approach mirrors that of other research on conspiracy theories emerging and circulating on social media [19][20]. As part of their initial collection process, the UCLA library confirmed that research use of the materials was in accordance with the terms of service of the sites. In addition, as part of their data preparation process, the library ensured that the collection was free from PII. After accessing this data through the UCLA library, we removed images, urls, videos, advertisements, and non-English text strings to create our research corpus, pizzagate.txt. To the best of our knowledge and to the best of the knowledge of the library, neither our corpus nor the library data contains data from private discussions, private chat rooms, or any other sources with restrictions on access for public use or that may violate the terms of our IRB exemption.

For Bridgegate, we relied on an archive of news reports developed by the UCLA library from a series of sources focusing on the northern part of New Jersey. This collection is also available through the UCLA library's archive site. The seed articles for the initial collection were either tagged or otherwise directly categorized as being about the closure of the lanes on the George Washington Bridge, and additional articles were indexed based on that initial seeding. We subsequently cleaned this collection to remove images, urls, videos, advertisements, and non-English text strings to create our research corpus, bridgegate.txt. All of our data can be accessed through a UCLA dataverse repository.

In its broadest outline, Pizzagate was "uncovered" by conspiracy theorists making use of the Wikileaks dump of emails hacked from the DNC servers, particularly those of John Podesta, who had served as the campaign manager for Hillary Clinton's unsuccessful run for the presidency in 2016. Through creative interpretations of these emails, conspiracy theorists alleged that they had discovered Hillary Clinton's involvement in a child sex trafficking ring being run out of the basement of a Washington DC pizza parlor, "Comet Ping Pong". The conspiracy theory took root with a series of tweets in early November 2016, with the first appearance of the #Pizzagate Twitter hashtag on November 6, the day before the US presidential election [26]. Discussions of the conspiracy theory tapered off, as measured by activity on Twitter, in December 2016, around the time that Welch was apprehended with his gun outside of the restaurant after surrendering to police [26]. Since then, Pizzagate has experienced a rebirth as part of the much larger QAnon conspiracy theory that began to develop in late October 2017.

By way of contrast, the Bridgegate conspiracy was discovered by investigative reporters to be a political payback operation launched by the inner circle of New Jersey Governor Chris Christie, making use of their close alliances with highly placed officials in the Port Authority. The conspirators took aim at the Democratic mayor of Fort Lee, New Jersey, Mark Sokolich, who had refused to endorse the governor in his reelection bid. Christie's assistants conspired with members of the Port Authority to close several toll lanes leading to the George Washington Bridge, thereby causing catastrophic traffic jams that lasted for a week in early September 2013. When asked, these people said that the lane closures were part of a traffic study. A formal investigation into the decision to close the lanes was launched in 2014 and, during the ensuing five years, the overall contours of the conspiracy were revealed and various actors were indicted, tried and sentenced to prison. In late 2019, a petition filed by several of the conspirators was granted review by the U.S. Supreme Court, with initial oral arguments occurring in early 2020.

For Pizzagate, our data set consisted of 17,498 posts comprising 42,979 sentences, with an end date of February 2018. We used a similar end date for Bridgegate, and thus worked with an archive of 385 news reports comprising 20,433 sentences. Because of this end date, we missed the events of April and May 2019 based on the revelations of one of the main conspirators, Bridget Ann Kelley, subsequent to her sentencing for her role in the conspiracy. These revelations highlighted the role of an otherwise seemingly unimportant actant, Walter Timpone, and added several new relationship edges to the Bridgegate narrative framework. The fact that additional information related to an actual conspiracy emerged over a prolonged period of time (here, five and a half years) might be one of the tell-tale signs distinguishing a conspiracy from a conspiracy theory. For Pizzagate, despite the three year scope of this study, the number of actants in the narrative remained stable beginning one month after the data collection period began.

Although Pizzagate was accessible through r/pizzagate and v/pizzagate, and the Bridgegate conspiracy was reported and archived by newspapers covering New Jersey politics, our approach does not require pre-established data sets. While access to comprehensive data collections eliminates an initial step in the narrative framework discovery pipeline, we have demonstrated methods for determining active domains of discussion in any collection of internet resources based on topic modeling [35][61]. Although the selection of a target domain using this and similar approaches might result in overlooking posts related to a broader discussion, work on community formation suggests that people interested in a particular topic seek out forums where such topics are discussed and develop close knit communities [20][29][46][62]. The first step in the pipeline can be tuned to capture actants that may be of interest; the extent of a domain can be discovered from there. In earlier work, we implemented this approach, and showed how a hierarchical topic-modeling method reveals broad topics of discussion in a large social media space that we identify as knowledge domains [35]. Posts, discussions and articles related to those knowledge domains can then be selected to constitute the study corpus. Cleaning the data results in a machine actionable corpus similar to those we developed for Pizzagate and Bridgegate. There are many other approaches that can be applied to the selection of target corpora from larger social media domains, although topic modeling has been to used to great effect in the context of social media [19][30][63].

## Methods

### A graphical narrative model for generation of posts and story fragments

We propose a generative network model, in which **actants** (people, places and objects) are the nodes, and **the relationships between pairs and groups of actants** are the edges or hyper-edges. These edges/hyper-edges are labeled with the nature of the observed relationships (for example, based on actions or attributes), the context of the relationships, and their likelihoods. We note that certain situations are better captured when they are represented as hyper-edges, involving multiple actants. Consider, for example, the verb/action "used" in the following sentence: "Podesta used the restaurant, Comet Pizza, to hide a ring for trafficking in children." In Semantic Role Labeling (SRL) [64][65] parlance, the verb "used" has at least three arguments or semantic slots: "X = Podesta = who uses" "Y = Comet Pizza = what is used by X", and"Z = Hiding a ring for trafficking in children = what X uses Y for." Thus a hyper-edge, connecting the actant nodes "Podesta," "Comet Pizza," and "Ring for trafficking in children" via the coupled semantic roles would be a sufficient representation. This hyper-edge can also be represented by a set of three pairwise relationships that are coupled: 1) ("Podesta", used, "the restaurant, Comet Pizza"); 2) ("Podesta", hid, "ring for trafficking in children"); and 3) ("Comet Pizza", hosted, "ring for trafficking in children"). For the remainder of this paper we assume that the networks only have pair-wise edges, and any multi-actant hyper-edge has been decomposed into a constituent set of coupled pairwise edges.

*This approach is **not** an attempt to redefine and build a semantic network* (in the usual sense), or an entity relationship network, such as Google's Knowledge Graph. In such networks, the actant categories or types are usually predefined, such as persons, organizations, and places. Similarly, different attributes and relationships among the actants are usually chosen from a predefined attribute list. For example, semi-automated databases will have a node entry for "Hillary Clinton" along with several relationship edges with other nodes such as, "(lives in), (America)", "(is a), (Politician)", and "(a member of), (the Democratic Party)", where the first argument is a relationship label and the second argument is another actant node. In this example "America", "Politician", and "The Democratic Party" are other nodes in the network. We make use of publicly available software platforms, such as *Flair* [66], to recognize named entities and their attributes, which helps us determine the various categories or knowledge domains to which the actants belong.

Our graphical models, by way of contrast, are primarily aimed at capturing actants and the interactant relationships that emerge under specific conditions and are driven by an underlying narrative framework. They are particularly suited for representing story and narrative dynamics where the overarching structure does not vary much, but the specific instances of the actants, their roles, and their relationships vary significantly based on the circumstances. For example, an "(arg1, relationship, arg2)" of the kind "(Hillary Clinton) (runs) (a covert child trafficking ring)" will not be included in any usual semantic network *a priori* (although it might get incorporated at a much later date, once the narrative has played out). Reporting that a public figure has "skeletons in the closet" is a common narrative trope, irrespective of whether it is true or not. Consequently, politicians and other public figures are monitored constantly by the press and other societal institutions that are keenly interested in discovering instances of the abuse of power or other criminal activities, and the "corrupt politician" is a well-known archetype. In the domain of politics, what varies are the identities of the actants, the nature of the crimes committed, and the motivations for committing those crimes or covering up the evidence. This means that the specifics of a "corrupt politician" narrative needs to be pieced together as pieces of information (whether credible or not) come to light.

Our computational approach to modeling story dynamics is to assume that the stories (and partial stories) are generated by an underlying domain-dependent structured model, where observed data is used to fill in the parameters of the model, as shown in Fig 1.

Formally, the dynamics of a particular narrative are characterized by an underlying set of $r$ relationships, $\mathcal{R} = \{R_1, R_2, \ldots, R_r\}$, and $k$ contexts, $\mathcal{C} = \{C_1, C_2, \ldots, C_r\}$. *These are model parameters that are either given **a priori** or estimated from the data.* A context $C_i$ is a hidden parameter or, to borrow a physics concept, the 'phase' of the underlying system, which defines the particular environment in which actants operate. It expresses itself in the distributions of the relationships among the actants, and is captured by a labeled and weighted network $G_{C_i}(V_{C_i}, E_{C_i})$. Here, $V_{C_i} = \{A_1, A_2, \ldots, A_n\}$, where each $A_j$ is an actant, and has associated with it a context specific probability or weight $p_{C_i}(A_j)$ that determines the actant's likelihood of participating in the given context. The edge set $E_{C_i}$ consists of $m_{C_i}$ ordered pairs $e_{(C_i,j)} = (A_{j_1}, A_{j_2})$, where each such pair is labeled with a distribution over the relationship set $\mathcal{R}$, $D_{(C_i,j)}(\mathcal{R})$.

Relationships are represented by categories of words (most often verbs) grouped together, where each category is comprised of verbs that imply a similar relationship. Therefore, the

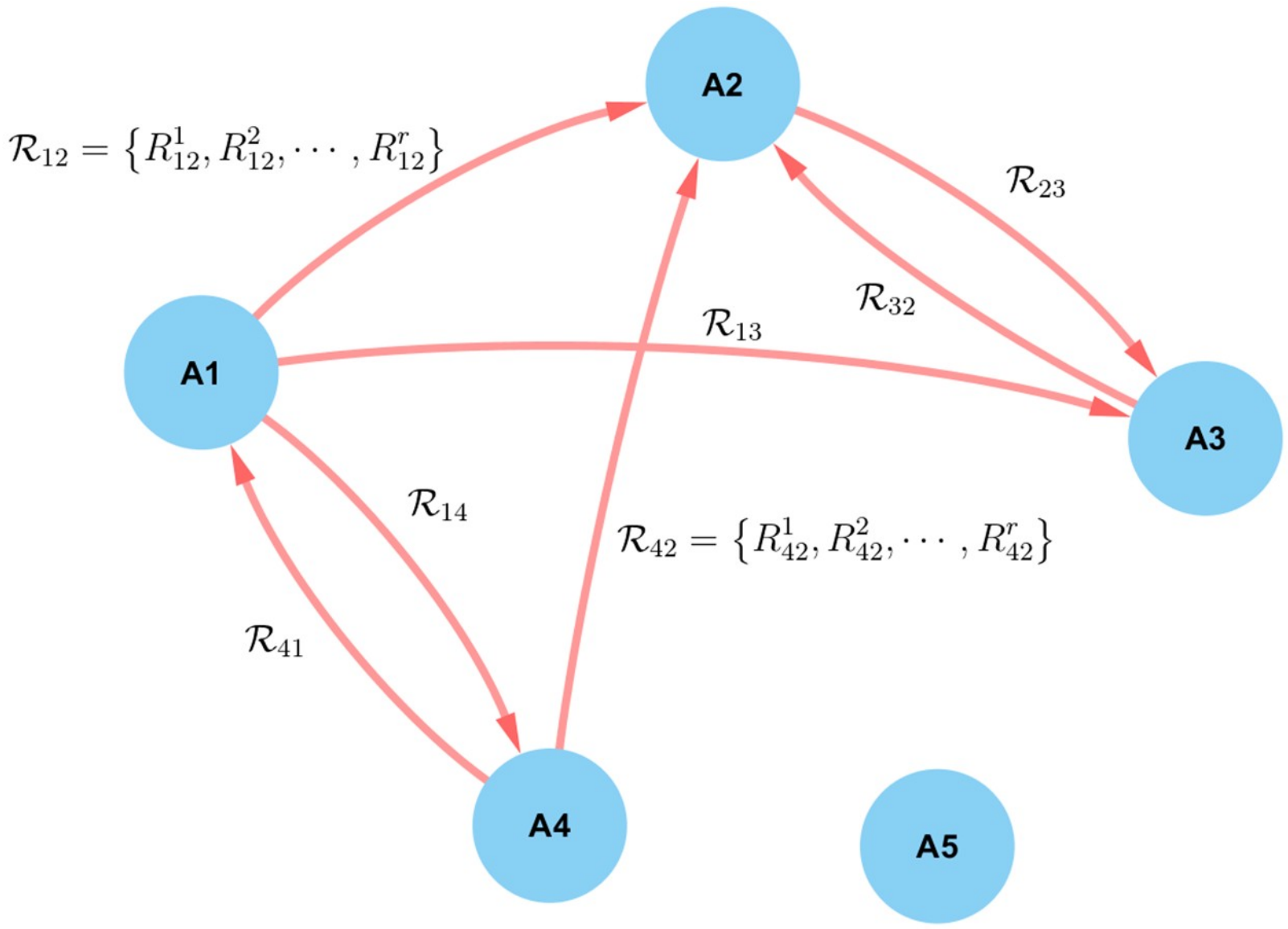

**Fig 1. A graphical model of narratives.** For a story with a set of actants $A_1, \ldots, A_n$, the narrative can be divided into a set of contexts. In each context, the story is summarized as a set of interactions (relationships) between actants as shown in the figure. Therefore, an edge between actants A1 and A2 carries a set of relationships $\mathcal{R}_{12} = \{R^1_{12}, R^2_{12}, \ldots, R^r_{12}\}$ that exist between the two actants, and the significance of each relationship in this context. It is important to note that relationships come from not only verbs, but also other syntactic structures in the text that imply relationships.

*(Abuse)* relationship is realized in sentences by a set of domain-specific verbs, including *(abuse, molest, rape, trafficking in)*, which connect the actants "John Podesta", "Hillary Clinton" and "James Alefantis" with the actant "Children" in the Pizzagate conspiracy theory.

## Learning narrative structure from large scale and unstructured text data

Our methodology is predicated on the underlying structure of the narrative framework that captures how a storytelling instance emerges via a collective negotiation process. Each post to a forum describes relationships among only a subset of actants (which are not yet known to our automated algorithms). The process of generating a social media post or a story fragment is shown in Fig 2. A person (user) first picks a context $C_i$ and then samples the network $G_{C_i}(V_{C_i}, E_{C_i})$ by drawing a set of actants according to the node distributions, $p_{C_i}(A_j)$. Then the user draws for relationships from among the associated distributions $D_{(C_i,j)}(\mathcal{R})$. The user then composes the post according to these outcomes by choosing the proper words and syntax, in particular nouns or noun phrases for the actants, and the associated verb phrases (or other syntactical constructs) for the relationships.

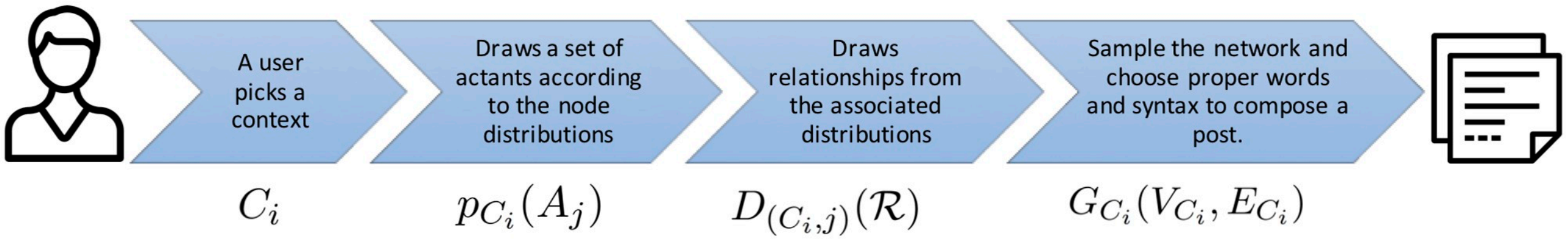

**Fig 2. Modeling the steps a user takes to generate a social media post or a story fragment for a given domain.**

This initial step is reminiscent of the experimental design of Raab et al. [14], who asked participants in their study (n = 30) to select and arrange phrases from a deck of cards containing various statements pertaining to the terrorist attacks of 9/11. These cards were arranged into subsets representing varying degrees of conspiratorial thinking. The resulting narratives were then evaluated for plausibility.

In our model, as information such as evidence or supposed "evidence" (in the case of a conspiracy theory) come to the fore, others may join the conversation and share their own information in terms of new or existing actants that either (i) affirm or contradict information in other posts, or (ii) bring in new information revealed either through a judicial or journalistic investigation process, or via the collective intent of an endemic population who see evidence and connections that others have apparently missed. The overall narrative is thus distributed across a series of social media posts and reports in different publications.

From a machine learning perspective, given such a generative process, we need to estimate all the hidden parameters of the model, including the actants, the set of relationships, and the edges and their labels. In other words, we have to jointly estimate all the parameters of the different layers of the model.

**Joint estimation of actants, contexts, and relationships.** We assume that the given corpus is a sample syntactic output of our graphical generative model. The underlying sets of actants, their semantic relationships and the contexts that determine different groups of relationships among the same actants are unknown. Thus, we need a formal data-driven function/measure to characterize these sets so that they can be estimated from the text corpus.

A functional model for actants can be described as follows: *An Actant is a set of Noun Phrases (e.g., named entities and head words in a parse tree) that play very similar semantic roles in the corpus.* The semantic role of a noun phrase is measured by the semantic similarity of the words and phrases around it in the parse tree. For example, (i) phrases such as "Clinton", "Hillary", "Hillary Clinton" form one actant category because of their high frequency, both as individual "head" words, and as co-occurring words in noun-phrases. As per our intuitive definition of an actant, because they are part of the same arguments in syntactic relationships, they have similar semantic roles; (ii) phrases such as "Supporter of Clinton", "Clinton follower" and "Clinton insiders" form a distinct semantic context because of the close semantic similarity of the words, Supporter, Follower, and Insider; (iii) phrases such as "Clinton Foundation", "Clinton Foundation Fundraising", "Clinton Donor" and "Clinton Foundation Contributions" form yet another distinct actant context because of the semantic similarities of the words Foundation, Fundraising, Donor, and Contributions. These examples guide not only the automatic determination of actants, but also reveal that the actants themselves have a hierarchical structure based on the different semantic and contextual roles they play. The phrases in (i) dealing with the different contexts for the actant Hillary Clinton can be considered a super-actant or a **supernode**, and the phrases in (ii) and (iii) dealing with different facets and distinct roles that are associated with the actant, Hillary Clinton, can be considered sub-actants or **subnodes**. The subnodes are specific semantic contexts that directly relate to the supernode and are expected to have relationships that are semantically homogeneous with the rest of the actant groups.

Historically, the semantic and functional similarity of words has been difficult to compute. In the past, these similarities were manually cataloged in dictionaries, thesauruses, and manually created databases such as WordNet and VerbNet. Recent advances in data-driven methods of embedding words and phrases into a multidimensional vector space [67][68] such that their Euclidean distances have correlations with their semantic similarity have made it possible to assign a quantitative measure to the similarity metric. The embeddings of syntactic argument phrases can be clustered with each cluster representing a separate actant. As we demonstrate in our results, this procedure of clustering embeddings of relationship phrases nearly automates the process of jointly estimating the actants and their attendant hierarchies.

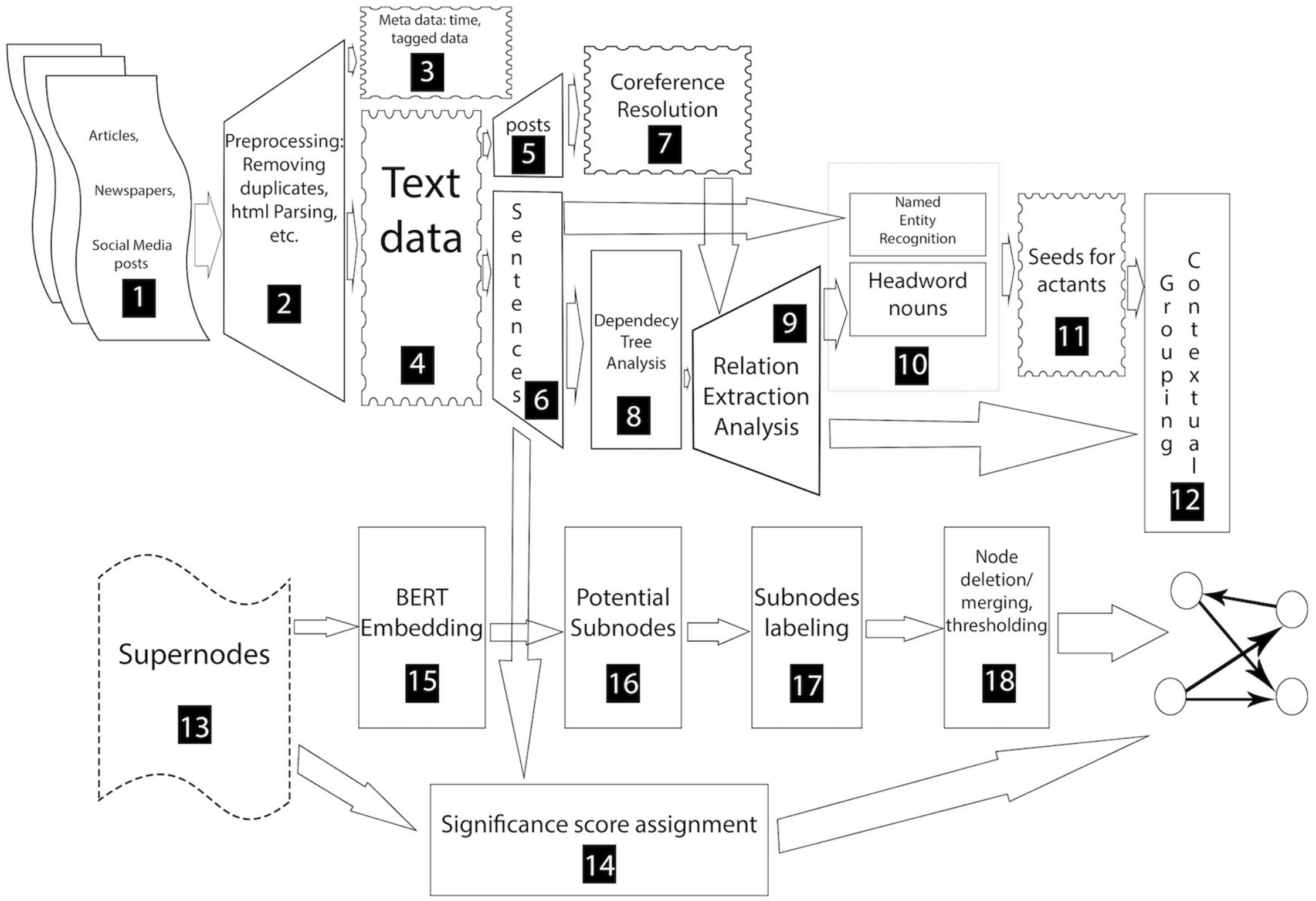

**Fig 3. Representation of the narrative framework discovery pipeline.** Most of these numbered blocks are described briefly in the main paper and in more detail in the S1 File.

Fig 3 provides a flowchart of the computational steps executed in our end-to-end pipeline. The salient computational steps are described below, and a more detailed description of each of the blocks is given in the S1 File.

*Syntax-Based Relationship Extractions (Blocks 7 and 8 in* Fig 3). Each sentence in the text corpus is processed to extract specific patterns of syntax relationship tuples in the form of ($arg_1$, $rel$, $arg_2$) where arg1 and arg2 are noun phrases, and rel is a verb or other type of phrase. Our relation extraction combines dependency tree and Semantic Role Labeling (SRL) [64] [65]. A similar, albeit more limited, approach to actant-relationship extraction is described by Samory and Mitra in their work on conspiracy theories [19]. In that work, their goal is to cluster semantically similar agent-action-target triplets, manually label the clusters, and align those labeled clusters with a manually curated topic model of a broader target corpus [19]. As opposed to limiting our extractions to agent-action-target triplets, we design a set of patterns (such as Subject-Verb-Object (SVO) and Subject-Verb-Preposition (SVP)) to mine extractions from dependency trees by using the NLTK package and various extensions [64][69][70][71] [72][73][74][75]; the patterns are based on extensions of Open Language Learning for Information Extraction (OLLIE) [76] and ClauseIE [77]. Second, we form extractions from SENNAs Semantic Role Labeling (SRL) model [70]. We combine dependency-based extraction techniques with SRL to increase the recall of our system. Then we apply cleaning and de-duplication techniques to select unique and high-precision extractions. A list of all the syntax relationship patterns, their definitions, and related examples are provided in the S1 File. The

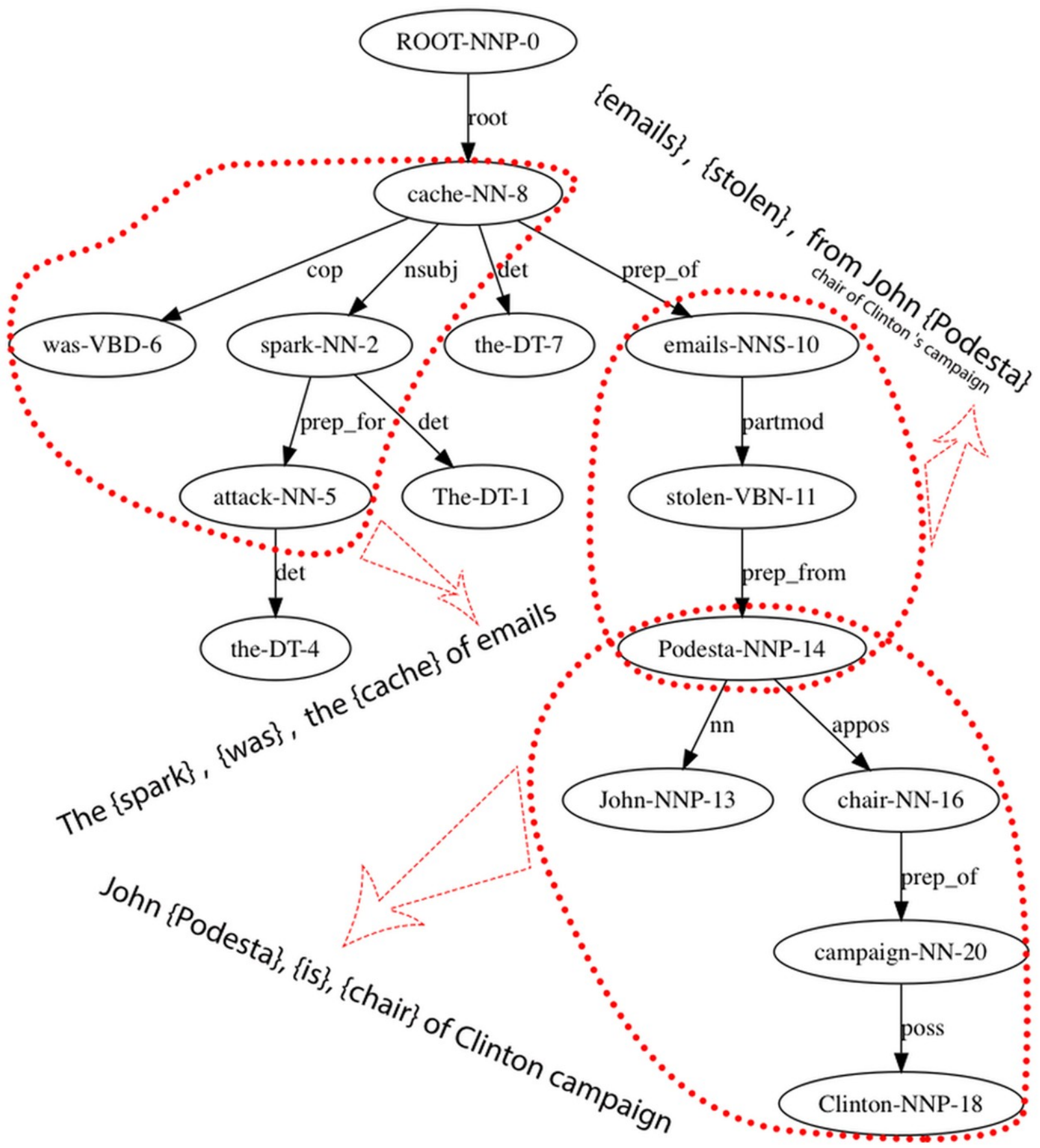

**Fig 4. An example of syntax-based relationship extraction patterns.** The sentence, *"The spark for the attack was the cache of e-mails stolen from John Podesta, chair of Clinton's campaign"* is analyzed to extract three relationship triples. These relationships are then aggregated across the entire corpus to create the final narrative network.

https://doi.org/10.1371/journal.pone.0233879.g004

sentence-level syntax relationship extraction task has been studied in work on Natural Language Processing [64][74][75][76][78][79] as well as in relation to discovery of actant-relationship models [19][80].

While extracting syntax relationships is an integral part of our framework, our work differs from previous syntax extraction work in one key aspect: *We take a holistic approach to extracting actants and their relationships*. As described previously, and in more detail in the following, the noun phrases arg1 and arg2 are aggregated across the entire corpus to group them into semantic categories or actants. This aggregation process (based on the generative model of narratives) also takes into account contextual differences, where the relationships between actants change in different situations. Such corpus-level structure cannot be inferred by simply extracting relationships seen in sentences. In our approach, syntax-based relationships, such as SVO (subject, verb, object), are tuned to capture story-specific syntactic forms of expressions. For example, to fit our generative model, we often break up three-way relationships into multiple pairwise relationships: a sentence, such as "The spark for the attack was the cache of e-mails stolen from John Podesta, chair of Clinton's campaign," is broken up into three pairwise relationships:: (The spark, was, cache of emails); (emails, stolen from, John Podesta); and (John Podesta, is, chair of Clinton campaign), as illustrated in Fig 4.

Because arg1 and arg2 could be pronouns, it is important that we determine to which nouns or noun-phrases these pronouns refer. Since pronouns often refer to nouns in preceding sentences, we use groups of sentences belonging to the same post as input to a co-reference tool (Stanford corenlp package as described in S1 File). We apply the output maps (pronouns resolved to nouns) to replace the resolved pronouns in the noun phrases, arg1 and arg2, with their corresponding nouns. As a result of this process, corresponding to block number 7, a major fraction of the pronouns are replaced by nouns. The input to this block is posts, and the output is used in block 9.

*Actant discovery (Blocks 10 through 18 in Fig 3)*. Formally, let $\mathcal{P}$ be the set of all relationship and noun phrases (i.e. all phrases, arg1, arg2, and rel occurring in any syntactic extraction (arg1, rel, arg2)). We define an embedding mapping $\mathcal{E} : \mathcal{P} \rightarrow \mathcal{R}^n$, that maps the set of phrases to a real vector of dimension $n$. Given any phrase, $\mathcal{P}_i \in \mathcal{P}$, $\mathcal{E}(\mathcal{P}_i) = \mathbf{y_i} \in \mathcal{R}^n$ (and without loss of generality we assume $||\mathbf{y_i}|| = 1$). Moreover, the mapping $\mathcal{E}$ is such that if $\mathcal{P}_i$ and $\mathcal{P}_j$ are semantically close phrases (i.e., they semantically mean almost the same even if they do not use the exact same words), then their corresponding embeddings must satisfy $||\mathbf{y_i} = \mathbf{y_j}|| \approx 0$. This requirement enables an unsupervised approach to actant determination: One can cluster the embedding vectors to obtain semantically close actant groups.

A direct approach to this problem might be to take all the noun phrases (i.e., arg1 and arg2) and get their embeddings using Word2Vec or Glove [67][68], or their more contextual embeddings using BERT [81], and cluster them using an algorithm such as $k$-means to obtain actant candidates. These clusters could then be further processed to merge very similar clusters to form a combined larger actant group, or to delete clusters that are not meaningful enough or are too heterogeneous (for example, as measured by the entropy of the word distributions in the phrases clustered together). This direct approach suffers from two major drawbacks: (i) The noun phrases, even after resolving pronouns/co-references, are dominated by high frequency pronouns, such as "they" "I" and "she", or not so meaningful online terminology, such as "URL". This ambiguity results in large clusters comprised of high-frequency but irrelevant actant groups, while more relevant actant groups get merged together to form heterogeneous clusters. (ii) The current embedding techniques tend to be flat (i.e., there is no inherent hierarchy in the vector space in which the words and phrases are embedded) and thus the example of the "Hillary Clinton" supernode and the subnodes related to "Clinton Foundation" and "Clinton Campaign" cannot be easily replicated.

The above observations motivated us to adopt a two-step process: (i) Contextual grouping of high frequency entities and concepts to create **supernodes**: First, we create a ranked list of named entities and concepts. Then we define a supernode as a context consisting of all the argument phrases that have a limited but unique and highly-correlated subset of the entities/concepts as substrings. In the Pizzagate corpus for example, we find all phrases with any of the following words {*Clinton*, *Hillary*, *HillaryClinton*} as one single supernode. Similarly, we find {*Pizza*, *CometPizza*, *Ping*, *Pong*} as the seed words for another supernode. Thus a supernode defines a general context, which can be further divided into subactants or subnodes as described below. (ii) Embedding vectors to cluster arguments in a supernode to create **subnodes**: Once we have defined meaningful contexts, we cluster the embeddings of the phrases belonging to a supernode to create subnodes.

*Determining supernodes (Blocks 10 through 13 in Fig 3)*. After retrieving syntax extractions from the corpus sentences, we generate and rank a list of entities, which is then used to form the seeds for potential actants. The ranking is based on the frequency of occurrences of the entities in the noun phrases arg1 and arg2. This ranking consists of both named entities as well as concepts such as "closures" and "email". For Named Entity Recognition (NER), we use the Flair framework [66], a character-level neural language model for contextualized string embeddings, along with the Flair pre-trained model. We limit the candidate actants to eight main types (see the S1 File for a list of the types). For concept discovery, we create a ranking of the frequent headwords in the noun phrases, arg1 and arg2. This method provides a second ranking of headwords including non-named entities. We then combine the two rankings, and rank each entity according to the summation of its frequency in the two lists (see the S1 File for examples of such entity lists). The list can be truncated to delete all nodes below a certain frequency threshold. The truncated list constitutes the original list of all entities/concepts to be considered for creating supernodes.

The subset of entities/concepts that define a supernode is computed in a hierarchical fashion: (**Step-0:**) The current entity/concept list is set equal to the original list. The maximum number of seed nodes in a supernode is set to $k$. (**Step-I:**) If the current list is empty, then Quit (supernode construction is complete). Otherwise, select the highest ranked entity/concept in the current list (in the first iteration, the entire original list is the current list). Let this entity be $E_1$. Add $E_1$ to the list of seed nodes for the new supernode, $S$. Remove $E_1$ from the current list. Set the seed-node list size, $|S| = 1$. (**Step-II:**) Find all phrases/arguments where any of the seed nodes in the set $S$ (i.e. the set representing the supernode under construction) appears as a sub-string, and let this be called $\mathcal{P}$. (**Step-III:**) Compute the most frequent entity/concept in the original list (other than the seed nodes already extracted) in $\mathcal{P}$. Let this be $E$. (**Step-IV:**) If $E$ has been processed before (i.e., it is no longer in the current list), then jump to **Step-VI**. (**Step-V:**) If $E$ is in the current list, then add it to the list of seed nodes, $S$. Remove it from the current list of entities/concepts. Increase the size count, $|S| = |S|+ 1$. If $|S| = k$ (where $k$ is the maximum size of the supernode seed list $S$), then go to Step-VI. Otherwise jump to **Step-II**. (**Step-VI:**) The current list of seed nodes, $S$, is the new supernode. Return to **Step-I** to start creating a new supernode.

*Subnode creation and labeling (Blocks 15 through 18 in* Fig 3). Each supernode represents a meaningful context, and is defined by its set of argument phrases. For each phrase we compute a BERT embedding [81] and cluster the embeddings of the phrases via $k$-means clustering, chosen for its simplicity, interpretability and suitability for our data [82][83]. Since supernodes have varying sizes (i.e. different supernodes have larger or smaller number of argument phrases), it is a computationally involved task to optimize $k$, the number of clusters for each supernode. In order to avoid such customization, we fix a single value of $k$ (for both Pizzagate and Bridgegate, we picked $k = 20$) for all supernodes and then delete insignificant clusters or merge two very similar clusters as follows: (i) **Deletion of small size clusters**: For each supernode, we plot the size distribution of the $k$ clusters, and we find that a certain percentage always has significantly smaller size than the average. Therefore, we define a threshold based on the ratio of the size of a cluster and the average size of the clusters for that supernode; all clusters with a ratio below this threshold are deleted. The rest of the larger clusters are processed as potential subnodes. (ii) **Merging of very similar clusters:** For each cluster, we generate a ranked list of the words that appear in the phrases that define the cluster. The ranking is based on a TF*IDF score, where TF is the frequency of the word/term in the phrases of the subnode, and IDF is the inverse of the number of paragraphs/posts that the word has appeared in the entire corpus. A list of $n$ (corpus dependent, $n = 2$ for Bridgegate and $n = 5$ for Pizzagate) top significant words from this list is then used to create a *label* for the cluster. For the particular implementation in this paper, we start with the first word in the ranked list, and then add the next word only if its score is greater than $\alpha^*(scoreofitspredecessor)$ for some corpus dependent $\alpha < 1$ (for Pizzagate we used $\alpha = 0.5$ and for Bridgegate $\alpha = 0.7$); if the next word is not significant then we stop. We also stop if we reach $n$ top words in this list of significant words. Thus, for each cluster we determine a label of at most $n$ representative words. Next we consider all the *k* clusters and merge all clusters with identical labels. *Each such merged cluster is now a subnode.*

*Contexts and context-dependent relationships*. For computational purposes, we define a particular context as the set of sentences where two actant categories as determined by noun phrases belonging to the same supernodes appear together in the same sentence. A context is characterized by the set of relationship phrases that have already been computed from these sentences. To further distill this set of relationship phrases and create a ranked order among them, we consider only the verbs in the relationship phrases because verbs are known to capture binary relationships in large-scale corpora [78]. The contexts defined by verbs have discriminatory power since they capture the different roles played by the same actants in different contexts.

*Computing significance scores for relationships (Block 14 in* Fig 3). In order to establish the significance of a relationship as summarized by their verb phrases for a particular pair of actants (i.e., a context), we employ the notion of conditional probability: A verb is contextually significant if:

$$P_{pair} = Prob[\text{verb}|\text{the sentence has both actants}]$$

$$P_{corpus} = Prob[\text{verb in any sentence in the corpus}]$$

$$P_{pair} >> P_{corpus}$$

Such a measure attenuates the effect of commonly occurring verbs such as “has”, “is”, and “are” (for which $P_{pair} \approx P_{corpus}$), while accentuating topical verbs that describe meaningful relationships between actants. Since there are many verbs involved in any context, we rank the relative significance of the different verbs via a scoring/weighting function $f(P_{pair}, P_{corpus})$, and then select the top ones as the verb set to characterize the context. Following on Bigi [84], we empirically tested various scoring functions, including TF*IDF style scoring functions, and discovered that a Kullback-Leibler (KL) divergence score produced the best results for this data as evaluated by three native English speakers. For any verb, the higher the KL score, the more significant that verb is to the pair.

To implement the above idea computationally, we stemmed the verbs. For every stemmed verb, $N_v$, we computed

$$P_{corpus}(v) = \frac{N_v}{N}$$

where $N_v$ is the number of times verb $v$ occurred in the corpus, and $N$ is the sum of the frequencies of all the verbs in the corpus. Then, for any given context, defined as the set of all sentences where the two actants co-occur, we computed

$$P_{pair}(v) = \frac{N_v(C)}{N(C)}$$

where $N_v(C)$is the number of times verb v occurred in the given context, and $N(C)$ is the sum of the frequencies of all the verbs in the context. Then we computed

$$ln\frac{P_{pair}(v)}{P_{corpus}(v)}$$

for all verbs v, and ranked them in decreasing order to obtain the set of top verbs that characterized the given context.

*Multi-scale narrative network generation*. The network defined by all the subnodes and their relationship edges, which are labeled by the most significant relationship phrases/verbs, is the final narrative framework or frameworks for a particular corpus. This network will tend to have a relatively large number of nodes and high edge density. The subnodes and super-nodes play different roles with varying importance. Meaningful sub-networks can be extracted by projecting various facets of the narrative network such as power-relationship networks, ego networks, super-node level networks, and networks comprising a target set of entities or actants; these projections, in turn, can be used to support multi-scale analysis of a complex narrative.

*Structural centrality of nodes and edges*. Various measures of centrality and importance can be computed for each of the nodes and edges in the network. Eigen-centrality or PageRank for nodes, and betweenness for edges are example measures. A set of central nodes in a narrative network can be defined as a set of minimal size whose removal breaks up the network into disjoint connected components. For example, as illustrated in Fig 9, the removal of the Wikileaks supernode and its edges in the Pizzagate narrative network breaks it up into disjoint connected components that define different domains that the actants inhabit. For the Bridgegate narrative network, no small size set of central nodes exists because the rich set of connections among the main actants existed well before the conspiracy to close the lanes on the George Washington Bridge.

## Community detection

Intuitively, a *community* in a network is a set of nodes that are more "densely" connected within the set than with nodes outside of that set. Given the nature of the inter-actant relationship network, such communities correspond to the subdomains of interaction. Partitioning any given network into an optimal number of clusters or communities is a well-known problem in graph theory that is computationally intractable (i.e. the problem is NP-complete) [85]. Several approximate algorithms have been developed that generate both disjoint and overlapping community partitioning, depending on the precise definition of the "density" of any candidate community [86]. An algorithm based on the modularity index measure [85] has been shown to provide good partitioning and various implementations of the algorithm are widely used across numerous fields [87]. A potential limitation of this algorithm is that, for each run, it returns disjoint communities that can vary over different runs (based on the random choice of seeds used for each run). In practice, when averaged over many runs of the algorithm, one finds that: (i) nodes that are strongly connected appear together in the same community over a majority of the runs; these nodes can be said to form the *core nodes* that define a stable community; and (ii) nodes that are more loosely connected with the core nodes and therefore change their community assignments; these nodes can be considered as ones that are overlapping or shared among different core-nodes defined communities. In the context of narrative networks, both sets of nodes provide significant information about the different core actant groups, and how these core groups interact via shared actants.

In order to discover this nuanced community structure, we develop an algorithm described below. In the first step, given a network $G(V, E)$ (where $V$ is the set of nodes, $N = |V|$ is the number of nodes, and $E$ is the set of edges), our goal is to determine $M$ (to be determined) core-defined disjoint communities, $C_j$ ($j \in \{1, \ldots, M\}$), such that $C_j(i) = 1$ if node $i$ belongs to community $j$, otherwise $C_j(i) = 0$, where $i \in \{1, \ldots, N\}$. Since the core nodes are not shared, $C_j^T C_k = 0$ for any two communities, $j \neq k$. To determine both $M$ and the communities $C_j$'s, we run the Louvain heuristic community detection algorithm (in NetworkX [88]) $T_{max}$ times. Next, a co-occurrence matrix, $A$, is defined, such that its element $A(i, j) = k$, if nodes $i$ and $j$ co-occur in the same community $k$ times over the $T_{max}$ runs of the algorithm ($0 \leq k \leq T_{max}$). We normalize this co-occurrence matrix by dividing every entry by $T_{max}$, so that $A(i, j)$ is the probability that nodes $i$ and $j$ co-occur in any given run. We next create a graph by defining an adjacency matrix, $G_c(i, j)$, where $G_c(i, j) = 1$ if $A(i, j) \geq P_{th_1} = 1 - \in$, where $\epsilon > 0$ is a small number. Every connected component with at least two nodes in this graph defines a core of a community, $C_j$. The number of non-trivial connected components (i.e., connected components with at least two nodes), $M$, is the number of communities. Note that, by construction, the cores are disjoint.

In the second step, we extend each core community $C_j$ by bringing in nodes that co-occur sufficiently many times with any node in $C_j$. That is, for every $k \notin C_j$, if there exists an $i \in C_j$ such that $A(i, k) \geq P_{th_2}$ (where $0 < P_{th_2} < P_{th_1}$), then $C_j(k) = 1$. Thus, the core nodes have strong connectivity, and the extended nodes share sufficiently strong connectivity. Note that after extension, the communities can overlap.

Finally, each community network is formed by the subgraph of the original network, $G(V, E)$, defined by the nodes in each $C_j$. Community co-occurrence frequency counts are retained for each node, since nodes with lower co-occurrence can provide information on various components of the graph that are not part of the core communities. We disregard nodes that have co-occurrence probability less or equal than $P_{th_2}$.

**Algorithm 1:** Community detection for a network, $G(V, E)$, with overlapping nodes

**Result:** $C_j(i)$, $F$<br>
$A_{k,l} = 0$<br>
**for** $i = 1{:}\ T_{max}$ **do**<br>
  Run community detection algorithm on $G$<br>
  **if** *nodes k,l in same community* **then**<br>
    $A_{k,l} = A_{k,l} + 1$<br>
  **end**<br>
**end**<br>
Normalize by $A = A/T_{max}$<br>
$A'_{k,l} = A_{k,l} \geq P_{th_1}$<br>
Form Graph $G_c$ defined by adjacency matrix $A'$<br>
$M$ = Number of Connected Components in $G_c$ with at least two nodes.<br>
**for** *Connected component* $C_j$ ($|C_j| \geq 2$) $G_c$ **do**<br>
  $C_j(i) = 0$<br>
  **if** $i$ $C_j$ **then**<br>
    $C_j(i) = 1$<br>
  **end**<br>
**end**<br>
**for** $i$, $k$ and $C_j$ **do**<br>
  **if** ($C_j(i)$ = = 1) *and* ($C_j(k)$ = = 0) and ($A_{i,k} \geq P_{th_2}$) **then**<br>
    $C_j(k) = 1$<br>
  **end**<br>
**end**<br>
For each $C_j$ construct a subgraph of $G(V, E)$ with nodes in $C_j$. $F$ is the union of all the community networks.

Once a community structure is determined for the narrative network (defined over all the subodes), we do further processing to discover the most frequently activated communities, actants and relationships. In particular, we filter for subnodes with a corpus frequency $\geq$ the average frequency count of subnodes in the corpus. The surviving subnodes are then grouped by actant supernodes. This step allows us to identify the central narrative framework. It also helps us identify less frequently activated communities and their constituent nodes, which may include components representing meta-narratives, unrelated conversations, or the

emergence—or to borrow a more fitting term from physics, *nucleations*—of other narrative frameworks.

## Visualization

Visualization of the narrative framework as a corpus-level graph and as subgraphs takes place in two steps, the first fully automatic, and the second with user supervision to present more easily read visualizations. The initial graph and subgraph visualizations are produced with NetworkX, where actants are imported as nodes, and all relationships are imported as directed edges; labels on those edges are based on phrases with the highest significance scores [89].

Various supervised visualizations allow for user input for layout and labeling. MuxViz is used to visualize the interconnections between domains for multi-domain narrative frameworks [90], while Oligrapher, Gimp and Photoshop are used for visualizing narrative frameworks with substantive edge labeling [91]. These latter edge-labeled graphs are inspired by the hand-drawn graphs mapping power relations by Lombardi [53]. Transformation of the node and edgelists between the automated pipeline and the required file formats for MuxViz and Oligrapher is done through a collection of simple scripts. Additional visualizations are generated in other graph visualization packages such as Cytoscape [92] and Gephi [93]. Final parameterization of the visualizations are determined by the user in the individual applications.

## Evaluation

We evaluate our results by comparing the narrative graph we learn to an expert labeled "gold standard" graph (as opposed to a ground truth graph). The lack of ground truth for machine learning work based on data derived from social and news media is a well-known problem [94]. As with oral narrative traditions where there is no "correct" version of a story (or ground truth), there is no canonical version of Pizzagate against which one can compare. For news stories, while journalistic accounts attempt to represent a ground truth, that "truth" is often contested [95]. In many cases, it is not until long after the news event is over that canonical summaries of the event are published; even then, there can be considerable debate concerning whether the news has been reported accurately, and whether the canonical summary is an accurate one (i.e. ground truth). For Bridgegate, that canonical summary has yet to be written, in part because several of the indicted co-conspirators are appealing their convictions, and in part because additional information continues to be reported. Given this lack of ground truth data, we use high quality, expert labeled data for evaluation [96].

For both Pizzagate and Bridgegate, we use the NY Times illustrations as the basis of our gold standard evaluation data [55][56]. It is generally accepted that the reporters and illustrators of the NY Times, as reflected by their status in the field of journalism, are capable of creating high quality, expert labeled data. Consequently, we consider their illustrations and accompanying explanatory articles as fulfilling reasonable criteria for external, expert generated validation data. Yet, while all of the nodes were labeled in these illustrations, the relationships between nodes were either poorly labeled (Pizzagate) or labeled in an inconsistent manner (Bridgegate).

To generate the gold standard expert annotations for the Pizzagate relationships, we proceeded in steps. First, we kept the labels from the eight labeled edges in the original illustration. Then we employed a standard three-person annotator setup to generate labels for the remaining edges: two independent expert annotators, native speakers of English with experience in journalism and political science and trained in narrative mark-up, provided their own relationship labels based on a reading of the article accompanying the Pizzagate illustration. Once they had completed their annotations, whenever they were in agreement, that label was used as the label on that edge. Whenever they were in disagreement, an equally qualified arbitrator decided on which label to use as the label for that edge [97][98][99]. We did not ask the annotators to add additional nodes or edges, although both of them decided independently to annotate the Edgar Welch episode described in the article, adding two additional nodes: "Edgar Welch" and "Police". The annotators also added three additional edges: "investigates" and "shoots" from Welch to "Comet Ping Pong", and "arrest" from Police to Welch.

Unlike the Pizzagate illustration, the NY Times Bridgegate illustration included labeled inter-actant relationships. These labels were not consistent and, along with relationships, also included actant descriptors (e.g. "top Cuomo appointee"), evaluative statements of relationships (e.g. "They weren't."), and speculative questions (e.g. "What prompted Kelly to send this email?"). To address this problem, we used the same three-person annotation team as described above to derive clear inter-actant relationship labels from the illustration. As the speculative questions included in the illustration were issues raised by the illustrators and not a part of the inter-actant relationship graph, we did not include them in our revised gold standard graph.

To determine the accuracy of our narrative framework graphs, we performed two evaluations, one to measure the accuracy of our actant extractions and aggregations, and one to measure the accuracy of our interactant relationships.

For actants, we calculated, given a threshold, whether the nodes represented in the hand-drawn illustrations were present or not in our extractions, and then whether they were present or not without the threshold. We also counted the actants that we discovered that were not in the hand-drawn illustrations. This last measure is important since the hand-drawn illustrations do not represent a ground truth, but rather serve as an expert summary based on human assessment of the reports of an event. It is possible that even expert summaries such as the NY Times illustrations do not include key actants; this was the case for Pizzagate, where Bill Clinton and the Clinton Foundation, frequently mentioned actants in the narrative framework developed on the Pizzagate forum, were missing in both the illustration and the accompanying article. We report the accuracy of our extractions for actants in Table 3.

To evaluate the accuracy of our relationship extractions, we developed an automated algorithm comparing our relationship phrases to gold standard relationships. For a set of relationships between entities $J_{A_1A_2}$, we aim to find a mapping $h_{A_1A_2} : J_{A_1A_2} \to C_{A_1A_2}$, where $C_{A_1A_2}$ is the gold standard set of relationships between those entities. This process is described as follows: Use the scoring function $f_{cos}(a, b)$ to compute the cosine similarity between $a$, $b$. A gold standard relationship phrase is mapped to an automatically extracted relationship phrase only if its embedding is close enough to be considered a match, here cosine $\geq 0.85$. This algorithm seeks to approximate a maximum likelihood estimation problem; $\mathcal{L}$ represents the cosine similarity $f_{cos}$ implemented with thresholds:

$$h_{A_1A_2}(j) = \underset{C \in C_{A_1A_2}}{\operatorname{argmax}} \ \mathcal{L}(C, j), \ \ \forall j \in J_{A_1A_2}. \tag{1}$$

The evaluations of these interactant relationships are presented in Table 4.

## Limitations

There are limitations with the current methodology that we hope to address in future research. As noted, the data can be noisy, particularly when social media posts are the primary source, as was the case for Pizzagate. This noise can create considerable problems for relationship extraction: a missing punctuation mark, for example, can significantly change the dependency tree structure and lead to erroneous extractions of both the arguments and the relationship phrases. Also, while pronoun resolution is needed and desirable to improve coverage (that is, to capture relationships amongst entities when they are expressed in terms of pronouns), it can also add noise by resolving pronouns to the wrong nouns. We designed several automated rules so that the pipeline errs on the side of caution. For example, if a relationship is extracted from a very long sentence, and the actants are far apart, we disregard the extraction. Similarly, if pronoun resolution substituted a pronoun with a long noun phrase, we disregarded that resolution. Although these rules decreased the potential coverage of relationships and the identification of actants, it gave us higher confidence that the extracted relationships were correct. Even with these stringent measures in place, we can get noisy syntactic relationships at the level of sentences. The aggregation (we do not include relationships without a certain number of repetitions) and sorting of relationships via their significance scores considerably improves the accuracy of the summary relationships. This process of de-noising both our syntactic and aggregate extractions is an ongoing research project.

Because of ambiguities in extractions and the noisiness of the data, actant aggregation is not always accurate. Again, our methods err on the side of caution and tend not to resolve all duplicate entities so as to avoid incorrectly resolving distinct entities into a single node. Finding clear aggregations for relationships is equally challenging, although we expect that refinements to context aware embedding methods will help this process considerably. Similarly, assignment of supernodes to particular domains can lead to ambiguities.

Because of the *ad hoc* nature of many of the online resources for studying conspiracy theories, it is difficult to extract consistent time data. This problem is exacerbated by two factors: inconsistent approaches to time stamping on the blogs and forums that include this information, and the common practice of participants re-posting or quoting from earlier posts. Dating and time stamping is not as significant a problem for newspaper articles, which are often the most readily available sources for studying actual conspiracies, even though many newspapers publish articles directly from news services, thereby introducing duplication.

Currently, some hand labeling of supernodes and relationships for clarity in visualizations is inevitable. The relationship labels, in particular, are neither as semantically rich nor grammatically correct as human generated labels. Nevertheless, the automatically generated relationship labels act as an informative starting point for human generated labels. Since we are focused on deriving the generative narrative framework or frameworks for a particular corpus, we do not address the type of high level abstractions discussed by Mohr et al., and Samory and Mitra in their work on narrative elements [19][80]. While those works rely on expert supervision for the very high-level abstractions of narrative elements found in their studies, it may be possible in future work to extend our pipeline to derive those abstractions automatically.

The use of hand-drawn expert illustrations as the gold standard against which we compare also has certain limitations. The NY Times illustrations do not, for instance, present a full rendering of the two complex target narratives. Instead, this comparison data presents a "minimal set" of the actants and myriad relationships that comprise the narratives. In that sense, they provide representations similar to those created by Bearman and Stovel who describe their approach as one that, "reduc[es] complex narratives to simpler images" [42]. Importantly, and for extended utility, our approach provides additional granularity, which supports the analysis of the context-dependent inter-actant relationships. In addition, our abstractions can be updated rapidly, unlike hand-drawn illustrations that require human intervention to render them.

Our selection of Pizzagate and Bridgegate for analysis was predicated on the existence of these external narrative framework graphs against which we could validate our results. We deliberately chose a conspiracy that was already in the process of being adjudicated in the courts (Bridgegate), and a conspiracy theory that had reached a level of stability as evidenced by the lack of growth in nodes or relationships for a period of more than a year (Pizzagate). Nevertheless, we recognize that, since we focused on two individual case studies, it is not possible to generalize our observations about certain narrative structural features as characteristics of each genre. It would be interesting, albeit beyond the scope of this paper, to apply our pipeline to many other corpora of conspiracies and conspiracy theories to test our hypotheses about narrative stability and the alignment of disparate domains that we propose in this paper.

There are several other limitations including the inability of our pipeline to process either foreign language corpora or rapidly changing data sources in real time. While the pipeline currently works only on English language materials, one can add NLP tools tuned to other languages to the pipeline. An eventual expansion of the pipeline would be the implementation of language detection and appropriate branches for the detected languages, thereby facilitating the use of multilingual corpora. Real-time analysis of data streams is currently not possible with our pipeline, although this could be a productive avenue for future research.

There are also certain limitations with the visualization and navigation of our results. Since the visualization of the network graphs relies on several different software packages, inconsistencies across users can result in different visualizations. Similarly, navigating user generated visualizations can be difficult. A future narrative network navigator will need to provide more straight forward visualization tools; such a "macroscopic" narrative network navigator, which is a focus of our current and future research, should also make it easier for users to explore the results at multiple levels of granularity from a single platform, rather than having to move between software packages.

Finally, we have not addressed the sequencing of events in the stories. Consequently, our narrative frameworks provide a snapshot view of the the conspiracy theory and the conspiracy in question, and do not include the third component of Greimas's model, namely the order in which inter-actant relationships are established. Determining the sequence of events may be difficult, although it would likely be a fruitful avenue for future research.

## Results

The joint estimation of the narrative framework network described in the Methods section relies initially on the relationship extractions. This process provides us with a ranked list of candidate entities used to seed the discovery of subnodes and supernodes, and a series of inter-actant relationships (Table 1). For each of the two corpora, we find a very large number of relationships of various types and patterns (Fig 5). After tokenizing and stemming the extracted headword lists, the resulting unsorted grouping provides a seed for the subnode lists and supernode lists. Once we take the union of the arguments with each of these terms, and determine the BERT embedding for each argument, *k*-means clustering (k = 20) results in a series of subnodes. After pruning and merging, we determine the supernodes and their corresponding subnodes for each narrative framework (Table 2; a full table is available in the S1 File).

To evaluate our actant discovery, we compare the actants discovered by our methods with those in the gold standard evaluation data. Even when we limit our actant list to those mentioned more than fifty times in the corpus, our methods provide complete coverage of the actants in the evaluation data. For Pizzagate, we provide comparisons with the illustration

**Table 1. Summary statistics for the extracted graphs from the two corpora.**

| | Supernodes | Subnodes | Rel Extractions | Labeled Rel | Avg Degree |
|---|---|---|---|---|---|
| Pizzagate | 24 | 88 | 749 | 438 | 36 |
| Bridgegate | 134 | 144 | 5855 | 928 | 72 |

https://doi.org/10.1371/journal.pone.0233879.t001

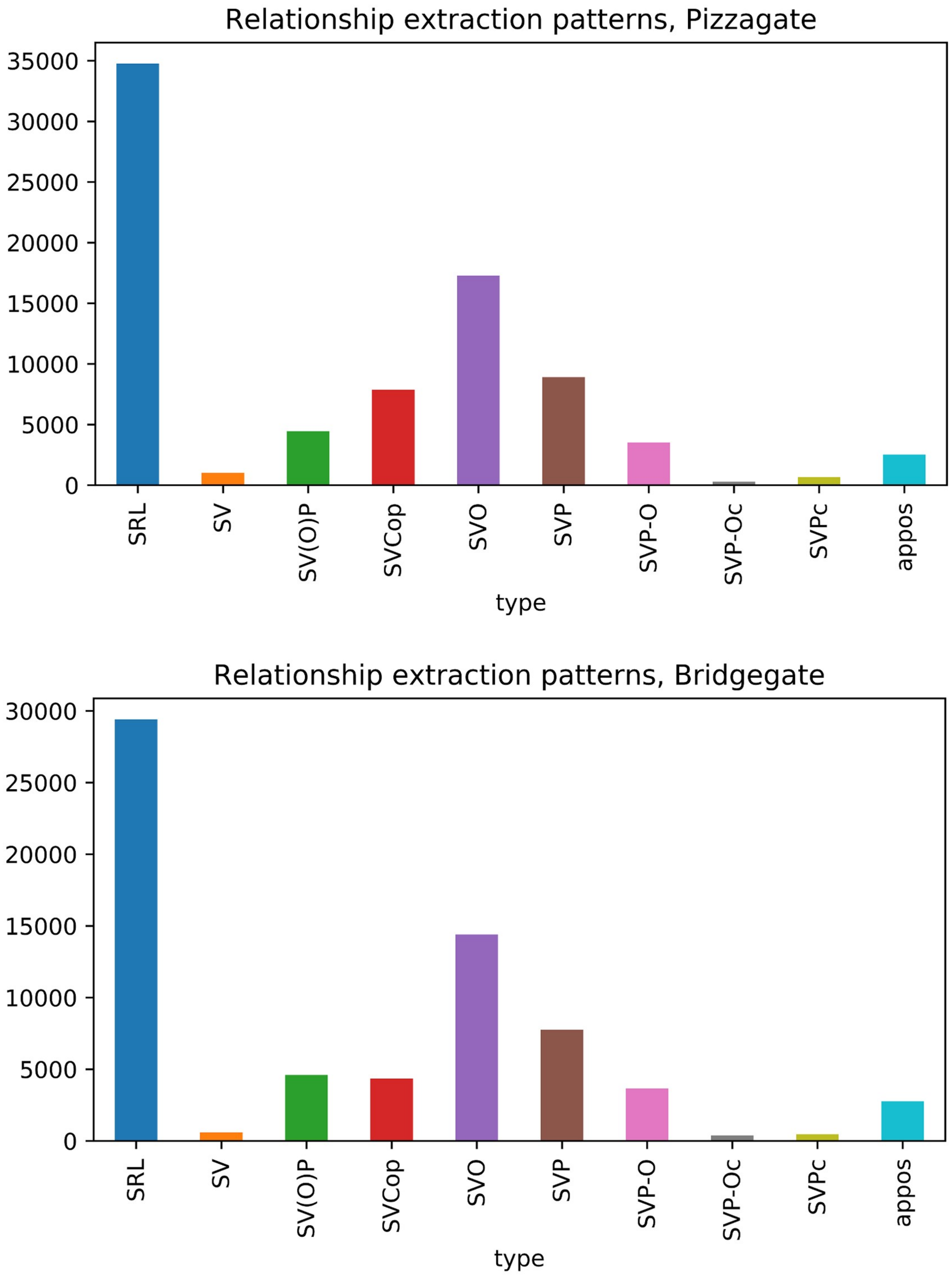

**Fig 5. Relationship extraction patterns.** Patterns by total number for A: Pizzagate (top) and for B: Bridgegate (bottom). For example, SVO is (nsubj, verb, obj), SRL is (A0, Verb, A1) and (A0, Verb, A2). A larger list can be found in S1 File.

**Table 2. A sample of the top 5 supernodes and subnodes for Pizzagate and Bridgegate.**

| Pizzagate | | Bridgegate | |
|---|---|---|---|
| *Supernodes* | *Subnodes sample* | *Supernodes* | *Subnodes sample* |
| [Podesta] | John Podesta,<br>Tony Podesta,<br>leaked Podesta email,<br>Podestas,<br>Podesta | ['christie',<br>'christi',<br>'christies',<br>'governor',<br>'chris',<br>'former'] | christie governor,<br>chris new jersey governor,<br>christie |
| ['pizza', 'comet',<br>'ping', 'pong'] | comet pizza,<br>comet pizza story,<br>ping pong comet,<br>comet,<br>ping pong review facebook | ['authority',<br>'author',<br>'authorizing',<br>'authorities',<br>'authors',<br>'authorization',<br>'port',<br>'executive'] | ['authority port',<br>'report authority port',<br>'executive director',<br>'baroni executive director',<br>'report',<br>'authority transportation'] |
| [alefantis] | James alefantis,<br>alefantis,<br>james alefantis instagram,<br>owner james alefantis | ['wildstein', 'david'] | ['wildstein',<br>'wildstein david',<br>'wildstein david executive former'] |
| [traffick] | child sex trafficking,<br>ring trafficking,<br>ring trafficking,<br>human pedophilia trafficking | ['lee',<br>'fort',<br>'mayor',<br>'sokolich'] | ['sokolich',<br>'fort lee',<br>'sokolich mark mayor',<br>' mayor effort sokolich',<br>'lee fort lane traffic'] |
| [child] | child,<br>child porn,<br>child trafficking | ['bridges',<br>'bridge',<br>'george',<br>'washington',<br>'lane'] | ['scandal bridge bridgegate',<br>'closure lane',<br>'george bridge washington closure lane',<br>'bridgegate',<br>'bridget kelly',<br>'closure gwb controversy lane',<br>'lane'] |

alone and with the expert labeled data, which includes the Edgar Welch meta-narrative (Table 3).

Our methods perform less well when actants are mentioned infrequently. The actant, "cannibalism", for instance, has a very low frequency mention in the Pizzagate corpus (4 mentions), and does not appear among the top ranked actants. Its inclusion in the NY Times illustration is visually arresting, however, which may in part explain why the illustrator chose to include it. By way of contrast, several highly ranked actants, including Bill Clinton and the Clinton foundation, do not appear in the NY Times illustration but are mentioned frequently in the Pizzagate discussions (Fig 6).

Similarly, some of the actants identified by the NY Times for Bridgegate are mentioned with relatively low frequency in our corpus. If, for example, we limit our actant list to only those mentioned more than 150 times (as opposed to 50 times), we miss five actants and their various relationships (Lori Grifa, Evan Ridley, Phillip Kwon, Paul Nunziato and Nicole Crifo).

**Table 3. Comparison of pipeline actant discovery with the gold standard evaluation data.**

| | New York Times | Pipeline Discovery | Matched > 50 | Matched anywhere |
|---|---|---|---|---|
| Pizzagate (illustration) | 21 | 88 | 20 | 21 |
| Pizzagate (expert) | 23 | 88 | 22 | 23 |
| Bridgegate | 36 | 144 | 36 | 36 |

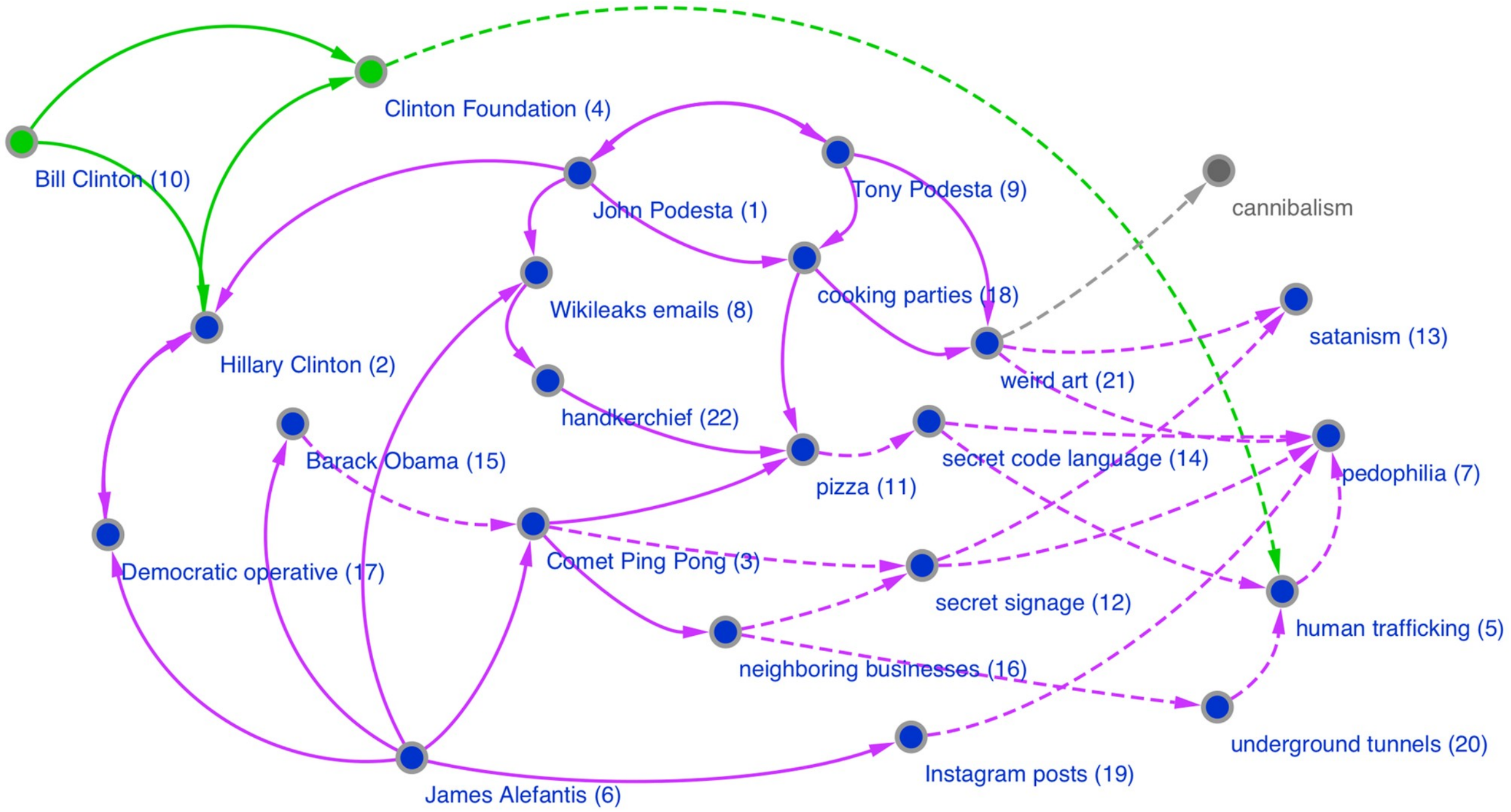

**Fig 6. Comparison of our results with the NY Times Pizzagate hand-drawn graph.** Edges and nodes that we do not discover in the top ranked actants through the pipeline are greyed out (cannibalism). Highly ranked edges and nodes that we discover not included in the NY Times illustration are in green (Bill Clinton and Clinton Foundation). We maintain the visual convention of dashed lines that the NY Times uses to identify relationships based on the interpretation by the conspiracy theorists of hidden knowledge. Immediately following the node label is the ranking of the actant as discovered by our pipeline.

These "misses", however, are replaced by actants such as Randy Mastro, a former federal prosecutor whose report exonerated Christie, Michael Critchlet, Bridget Anne Kelly's attorney, and Shawn Boburg, whose reporting broke the scandal, all of whom are central to the conspiracy and reporting on it.

Relationships between supernodes can be discovered by collapsing the subnode subgraphs, and labeling the edges between supernodes with the relationship with the highest relevance score over the subgraph edges (for example, Fig 7). A table summarizing the comparison of our relationship extractions and aggregations with the evaluation data lists the number of edges in the NY Times illustrations, the number of expert labeled edges in the gold standard corpus, and the overall number of automated aggregated relationship extractions from our pipeline, as well as the recall of our extractions against the gold standard relationships, the average cosine similarity score for matched edges, and the standard deviation for this measurement (Table 4).

## Discussion

Network representations of complex narratives are widely recognized as providing support for understanding narratives, including the actants, their various roles and the numerous interactant relationships of which they are comprised [42][43]. Our methods allow us to derive these narrative frameworks automatically, and present them as a network graph. Visual representations of narrative networks such as those included in the NY Times and the ones we generate have also become a part of the reporting on complex events including conspiracy theories such as Pizzagate and conspiracies such as Bridgegate.

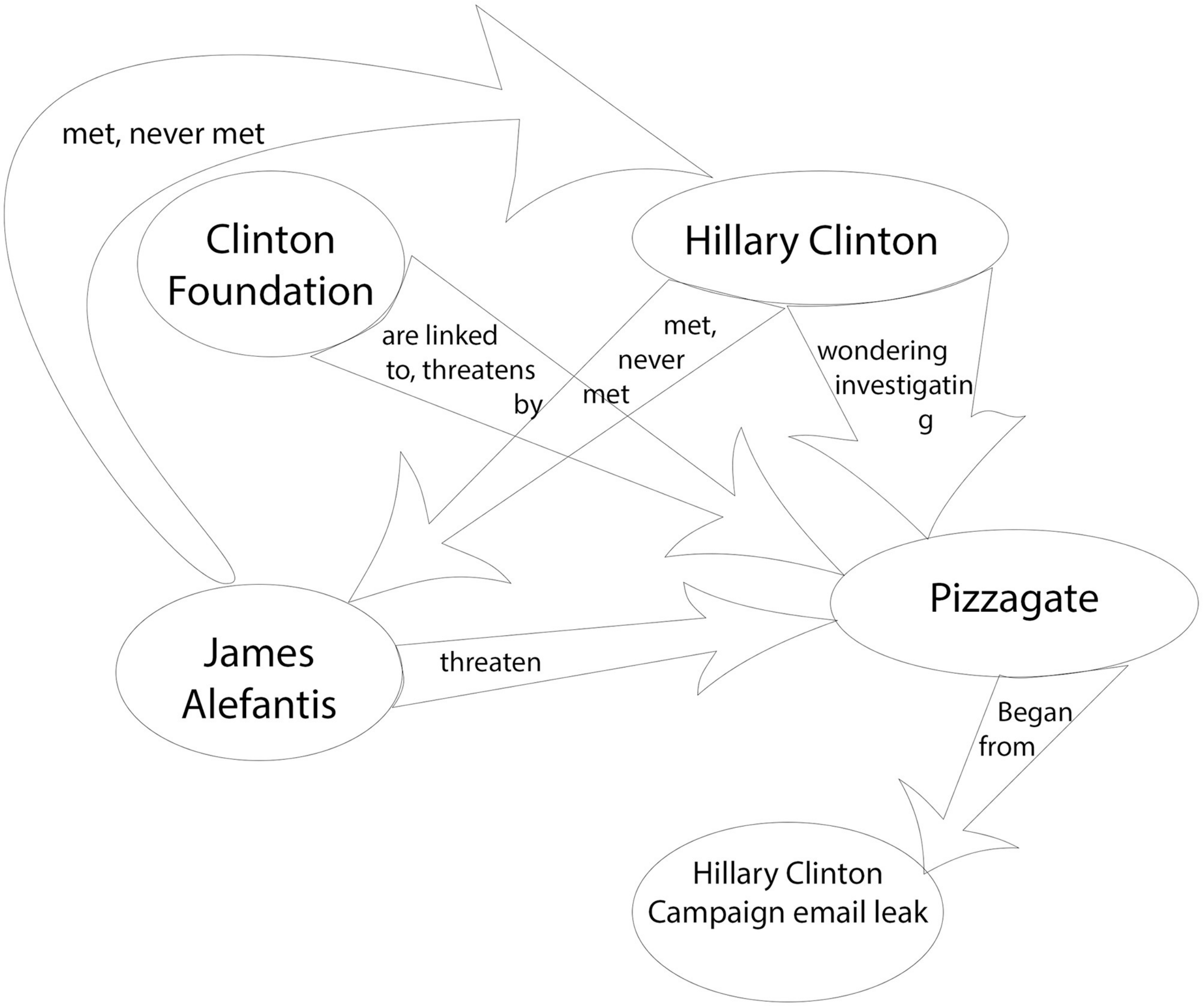

**Fig 7. A subnetwork of the Pizzagate narrative framework.** Some of the nodes are subnodes (e.g. "Clinton Foundation"), and others are supernodes (e.g. "Pizzagate"). Because we only pick the lead verbs for labeling edges, the contextual meaning of relationships becomes more clear when one considers the entire relationship phrase. For example, the relationship "began" connecting "Pizzagate" to "Hillary Clinton Campaign email. . .." derives from sentences such as, "*What has come to be known as Pizzagate began with the public release of Hillary Clinton campaign manager John Podesta's emails by WikiLeaks. . .*". Similarly the edge labeled "threaten" connecting "Alefantis" to the "Pizzagate" supernode is derived from sentences such as, "*James Alefantis threatens Pizzagate researcher. . ..*". Here the supernode, "Pizzagate" includes the entity "Pizzagate researcher," which appears as a subnode.

Running the community detection algorithm on Pizzagate reveals thirty-three core communities. After applying the thresholds on actant mention frequency and community co-occurrence as described in the Methods section, here $P_{th_1} = 0.7$ and $P_{th_2} = 0.4$, we discover a series of seven cores, corresponding to the main Pizzagate domains, as well as five nucleations of potentially emerging narrative frameworks, two meta-narrative cores, and two possibly unrelated cores (for calculation of these thresholds, see S1 File). A visualization of those communities that only include subnodes with mention frequencies greater than or equal to the corpus average mention of 265 reveals the distinct narrative framework underlying the Pizzagate corpus, where central cores have a large number of edges connecting them (Fig 8). Subsets of the supernodes define four domains that, outside of Pizzagate, would probably be less connected: (i) Democratic politics, where actants such as Hillary Clinton and Obama are dominant; (ii) the Podestas, with John Podesta as the major actant; (iii) casual dining, dominated by James Alefantis and Comet Ping Pong; and (iv) Child Sex Trafficking and Satanism, where actions such as child abuse and sex trafficking, and actants such as children and rituals are common. Subnodes in the self-referential meta-narrative, "Pizzagate", have many edges connecting them to these core domains, while the narrative nucleations and unrelated discussions do not. This lack of connection suggests that they are not central to the Pizzagate narrative and their corresponding mention counts further reveal that they are not discussed frequently.

**Table 4. Comparison of pipeline inter-actant relationship discovery with the NY Times and the gold standard corpora.**

| | NY Times illustration | Gold-standard corpus | Automated Extractions | Recall | Avg cos similarity | Std Dev |
|---|---|---|---|---|---|---|
| Pizzagate | 35 (27 unlabeled) | 38 | 749 | 83.7% | 0.95 | 0.048 |
| Bridgegate | 46 | 122 | 5855 | 82.9% | 0.89 | 0.0483 |

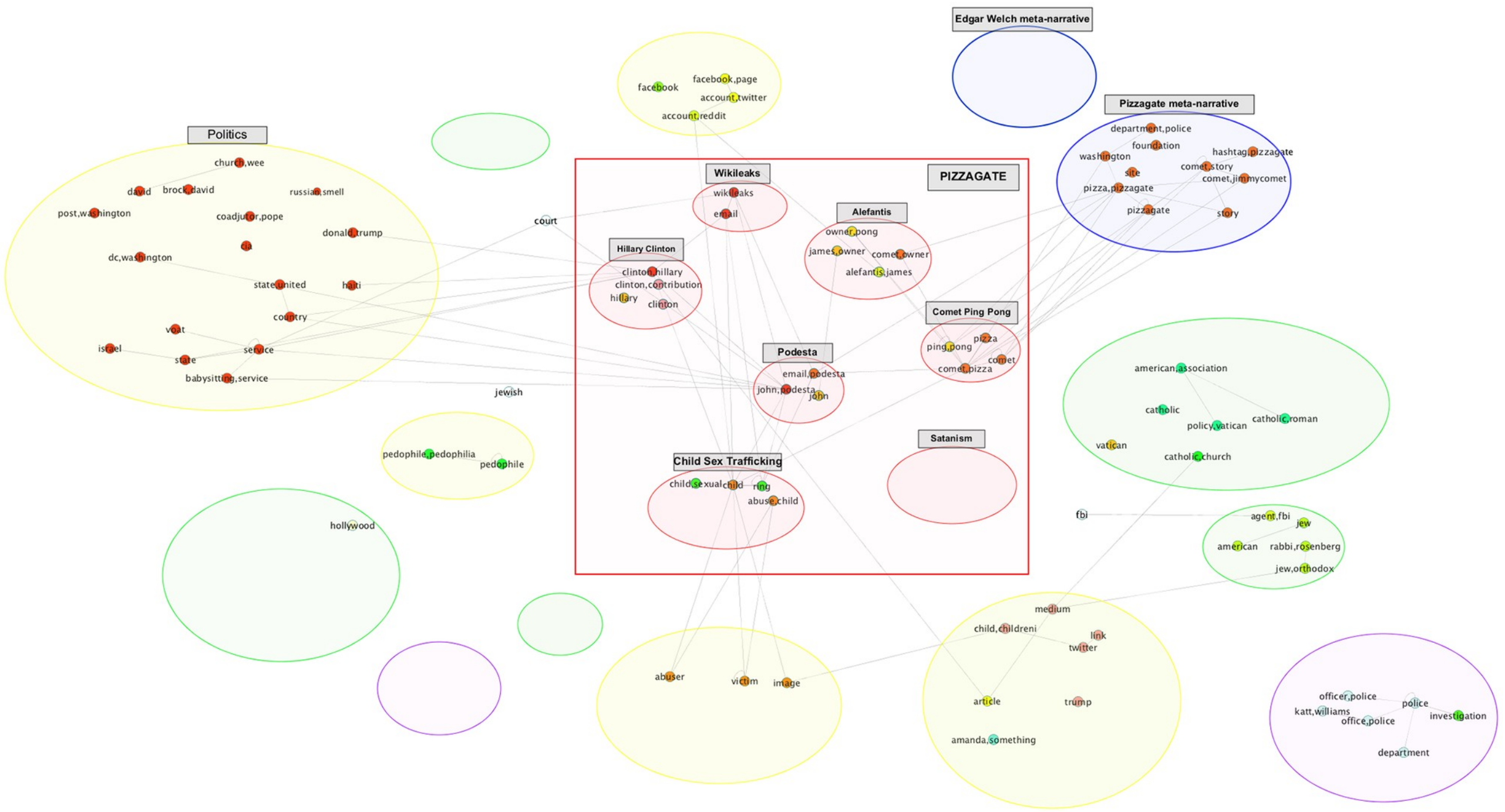

**Fig 8. Identification of the Pizzagate narrative framework from the Pizzagate corpus.** Subnodes with a mention frequency count < 265 and their edges are removed from the community-partitioned network obtained from Algorithm 1 (See Fig 10 for the network before filtering). Solid nodes are core nodes, while nodes without color, such as "fbi", are non-core nodes. Colors are based on the core nodes' assigned community, while all relationships are collapsed to a single edge. These core nodes have an assignment based on the $P_{th_1} = 0.7$ threshold, while open shared nodes have an assignment based on $P_{th_2} = 0.4$ threshold (see Algorithm 1). Pizzagate subnodes are concatenated into their supernodes, and are outlined in red, while the subnodes retain their community coloring. Contextual communities are shaded with yellow, metanarrative with blue, nucleations with green, and unrelated discussions with purple.

It is interesting to note that the Wikileaks domain, dominated by actants such as email and Wikileaks, provides the glue for the narrative framework. After eliminating the relationships generated by the Wikileaks subnodes, the connections between the other domains disappear, leaving them as a disjoint series of smaller connected components (Fig 9). This disjuncture only occurs when the links generated by the Wikileaks subnodes are eliminated.

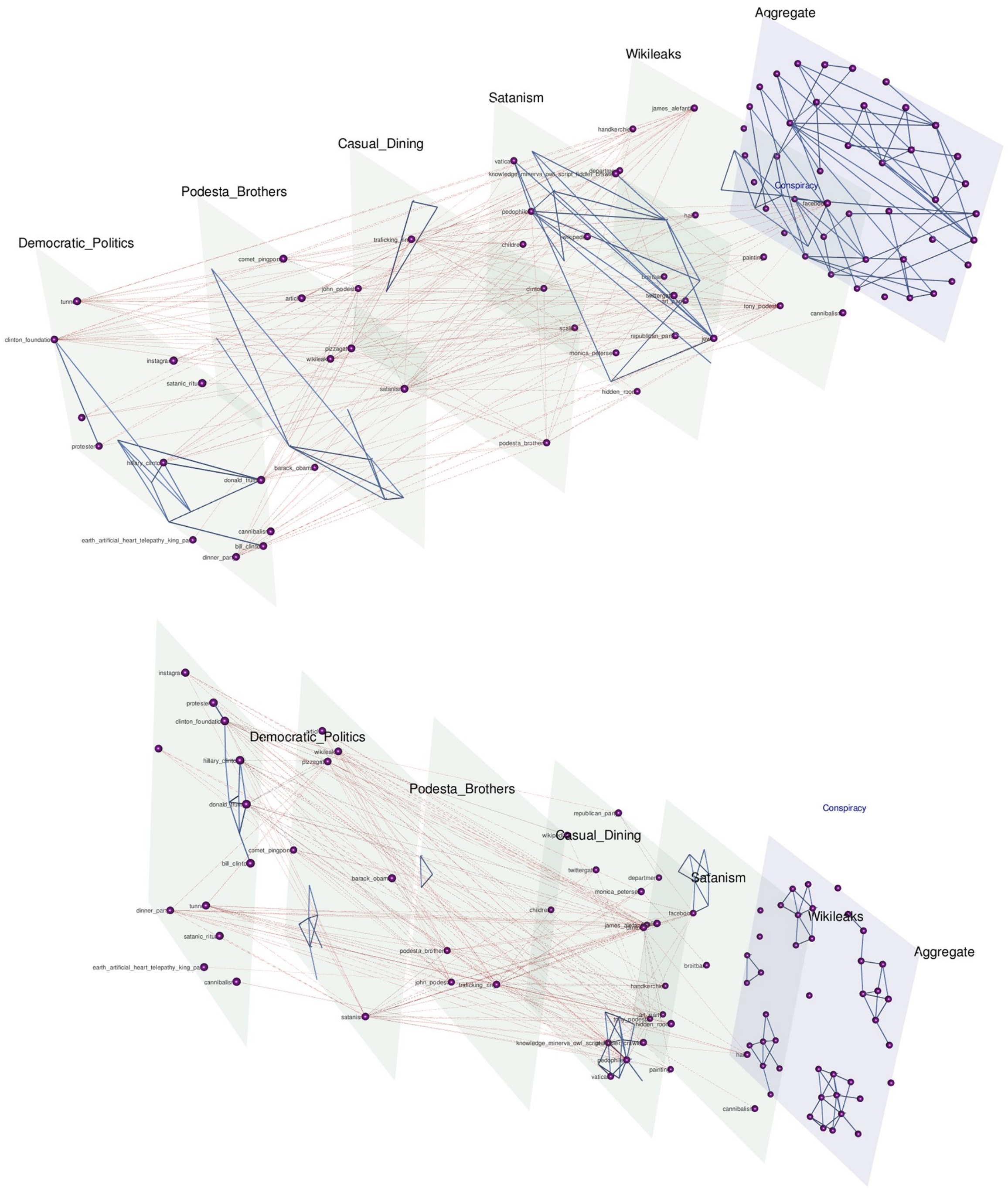

**Fig 9. A three dimensional visualization of the narrative framework for Pizzagate in terms of domains.** On the top, **A:** the graph with the inclusion of relationships generated by Wikileaks—the aggregate graph in blue shows a single large connected component. On the bottom, **B:** the graph with the Wikileaks relationships removed, shows on the aggregate level the remaining domains as disjoint components. In the Pizzagate conspiracy theory, the different domains have been causally linked via the single dubious source of the conspiracy theorists' interpretations of the leaked emails dumped by Wikileaks. No such keystone exists in the Bridgegate narrative Network.

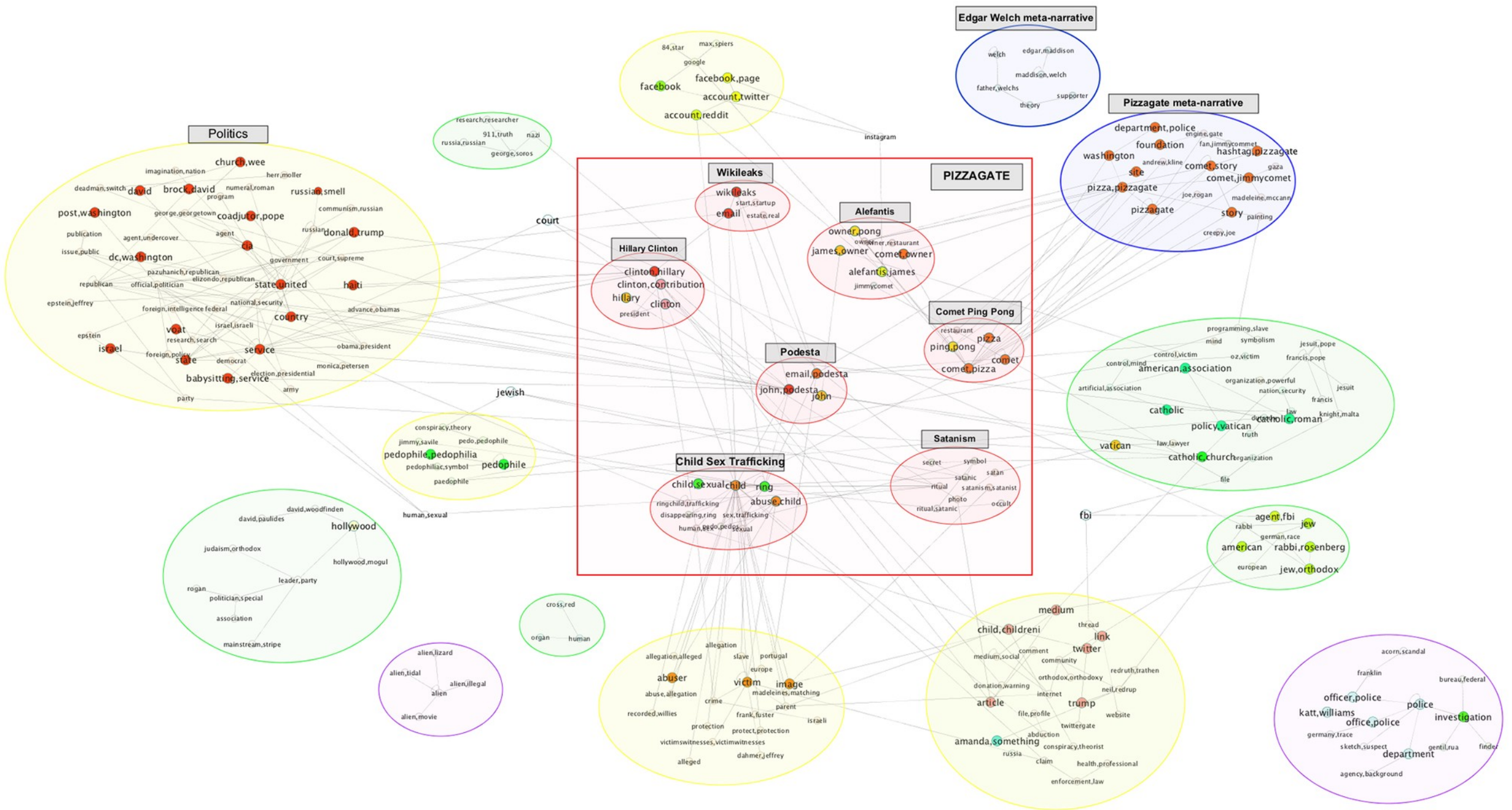

**Fig 10. Community detection on the overall Pizzagate corpus.** Subnodes are colored based on their assigned community, while all relationships between any two subnode actant nodes are collapsed to a single edge. Solid core nodes have an assignment based on the $P_{th_1} = 0.7$ threshold, while open shared nodes have an assignment based on $P_{th_2} = 0.4$ threshold (see Algorithm 1). Main Pizzagate supernodes are outlined in red, and include their subnodes colored by community. Meta-narrative frameworks are shaded with blue. Context groupings are shaded with yellow, while narrative framework nucleations are shaded with green. Unrelated discussions are circled in purple. The entire Pizzagate narrative framework is highlighted with a red box (see Fig 8 for a frequency-filtered version of this figure).

When we remove the mention frequency threshold of 265, the visualization is populated with the subnodes comprising the various communities (see Fig 10). Apart from the central Pizzagate narrative framework, an important meta-narrative component describing Edgar Welch's investigations of Comet Ping Pong appears. Although this meta-narrative is mentioned in the NY Times article on Pizzagate that accompanies the illustration, it was not included in the illustration [55]. Importantly, our detection of the Welch meta-narrative matches the annotations of the expert annotators as reported in the evaluation of our results above. A second meta-narrative focuses on Pizzagate as a topic of discussion, and includes references, for example, to the Pizzagate hashtag. Apart from these two meta-narrative communities, there are several other communities that we detect in the overall graph: (i) communities that provide background or support for the central Pizzagate narrative framework; (ii) communities that may represent nucleations of other narrative frameworks; and (iii) communities that are unrelated to the Pizzagate narrative framework.

Several of the communities providing background describe the internet itself and various social media platforms, thus presenting a general internet-based context for these discussions. Two additional background communities focus on general discussions about pedophilia, and various allegations against people such as the British entertainer and sexual abuser, Jimmy Savile. A final large background community focuses on American politics writ large, and provides the domain from which the various democratic operatives and Obama are drawn.

Other communities represent the beginnings of other narrative frameworks, which either represent indigenous nucleations of new narrative frameworks, or the intrusion of additional narrative frameworks from outside the target forum. One of these communities is redolent of anti-Semitic conspiracy theories related to Hollywood, which are common in other conspiracy theory forums, and may indicate the existence of conversations that propose links between Pizzagate and these broader conspiracy theories [19]. The rise of QAnon, which includes both anti-Semitic conspiracy theories and the Pizzagate conspiracy theory, suggests that this may be the case [100]. Another small nucleation relates to suspicions that the Red Cross traffics in human organs. Other components that may represent emerging narrative frameworks include a community focused on Rabbi Nuchem Rosenberg and his efforts to reveal child abuse in certain Orthodox Jewish communities, and a community that includes narrative components related to secret orders within the Catholic Church, including the Knights of Malta. One final nucleation presents a possible narrative about George Soros, 9/11, Russia and Nazis.

There are two unrelated communities–one focused on discussions of aliens and alien films, including the film *Alien*, while the other is related to discussions about police and FBI investigations of Acorn and Katt Williams. These last two communities reveal how our methods work even with very noisy data that may include conversations not immediately relevant to the discovery of the narrative framework(s) in the target corpus. It is important to note that all of these non-central components are comprised of actants and their relationships that have lower than average frequency mentions in the corpus.

For Bridgegate, we discover a much simpler community structure, with a single giant connected component of 386 nodes. The community detection algorithm finds twenty-three communities, but only three of them have 20 or more nodes, with a mean size of 6.65 and a median of 3 for the remaining communities. This result is not surprising given that all of the actants in the Bridgegate conspiracy come from a single domain, namely that of New Jersey politics. Consequently, the narrative framework is not stitched together through the alignment of otherwise weakly connected domains, but rather is fully situated in a single domain. Similarly, there is no information source, such as Wikileaks, on which the framework depends to maintain its status as a single connected component. Even the deletion of a fairly important actant, such as Bridget Kelley along with her relationships, does not lead to a series of disjoint subgraphs as was the case in Pizzagate when the Wikileaks associated nodes were deleted. Indeed, even if all of the Bridgegate actants' conspiracy-related relationships were deleted—as if the conspiracy had never happened—New Jersey politics (for better or worse) would continue to exist as a giant connected component.

In addition, a time series analysis of the Bridgegate data reveals that, unlike the Pizzagate data in which all of the actants emerged over the course of approximately one month, the cast of actants associated with Bridgegate took nearly six years to be fully described, with several spikes in the emergence of new actants related to the discovery of new aspects of the conspiracy and various court cases (Fig 11).

While community based graphs present a clear macro-scale overview of the main categories of actants for the narrative, and thereby offer the type of overview described by Lehnert and others [43], they elide the important meso- and micro-scale aspects of context dependent subnodes and interactant relationships fundamental to the macroscopic approach we propose here [52]. Unfortunately, including a complete set of labeled edges between subnode pairs in a single visualization is difficult [101]. To address this problem, for meso-scale and micro-scale analysis, we generate subgraphs, such as ego-networks, and their visualizations. Visualizing the ego-networks for any of the subnodes results in an image with clear labels on the nodes and edges, as shown with a selection of nodes for the Podesta subnode egonet (Fig 12).

To present these finer grained representations of the narrative framework, we create two different networks describing entity relationships for each of the subnodes. One network, which we label the "power network", includes only named individuals and their professional or personal relationships. An example of an actant/relationship pair in this network for Pizzagate is: "John Podesta is Hillary Clinton's campaign chief." A second network is derived from contextual and interaction based relationships, such as the "Podestas had dinner." Since each subnode represents a contextually sensitive use of the supernode category, we can visually represent these subnodes and relationships as subgraphs for each supernode. For example, the subgraph for the Podesta supernode from Pizzagate reveals a series of attributes of Podesta (e.g. Clinton campaign manager), and a series of context dependent relationships (e.g. having dinner with his brother) (Fig 13). In another example, a sub-selection of named individual nodes from the Bridgegate graph, including Bridget Anne Kelly, highlights the types of relationships between her and others in her ego network (Fig 14).

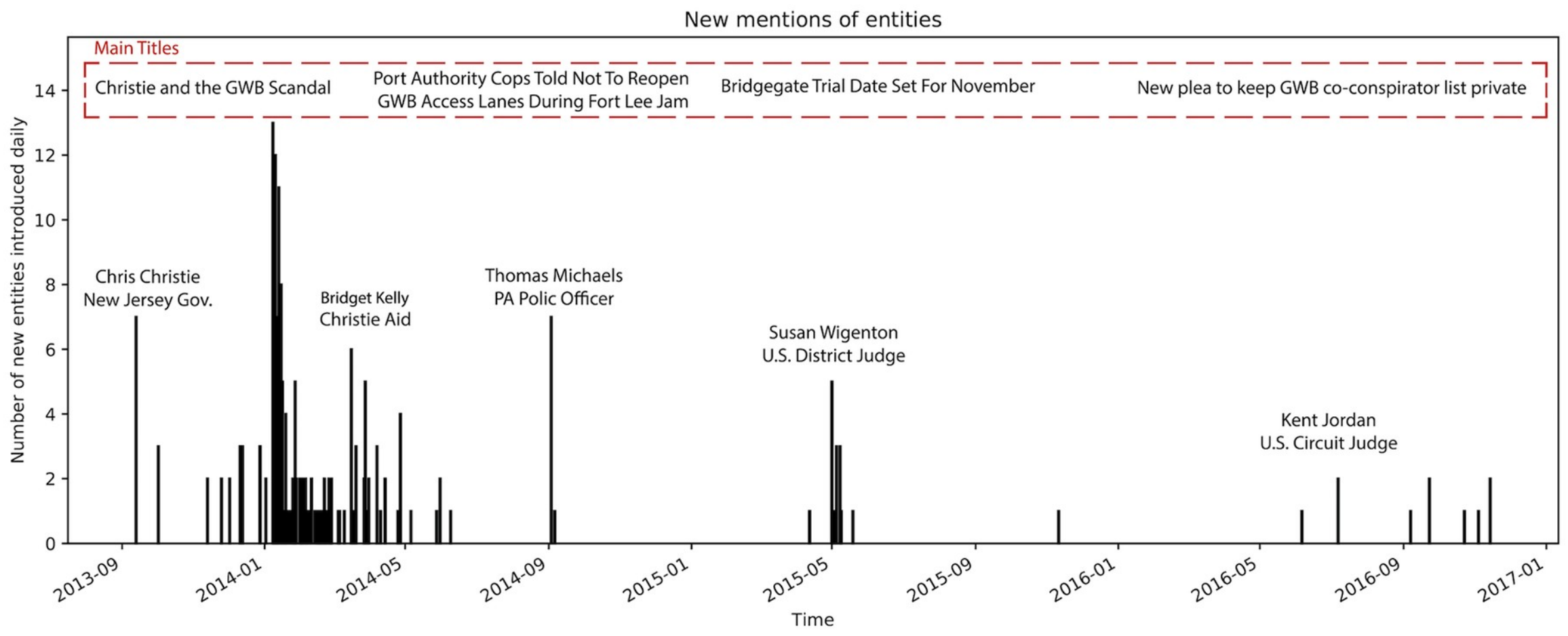

**Fig 11. Time series of the first mention of Bridgegate entities.** Starting with the events of September 2013.

Developing a complete understanding of the narrative framework proceeds in steps: the community level macro-visualization provides an overview of the main actants, the core communities, and the most frequently mentioned subnodes and their relationships. Drilling down to the meso- and micro-scales offers increasing detail. Navigating across the various subnode ego-networks, with their semantically rich edges between subnode pairs, provides a comprehensive understanding of the actants and their relationships. These meso- and micro-scale representations of the narrative framework provide access to significant detail not found in the summary graphs such as the illustrations from the NY Times. For example, while there is no link between John Podesta and Satanism in the NY Times graph, our subgraphs reveal that, according to the conspiracy theorists, he was a follower of Satanism. These subgraphs also reveal additional rich relationships missing from the NY Times graph, such as John Podesta's brother Tony's ownership of weird art that used coded language to promote pedophilia (Fig 15). For the Bridgegate conspiracy, an examination of the Chris Christie ego-network, for example, provides more nuanced relationships between the actants that populate both the NY Times illustration and our finer-grained narrative framework network (Fig 16).

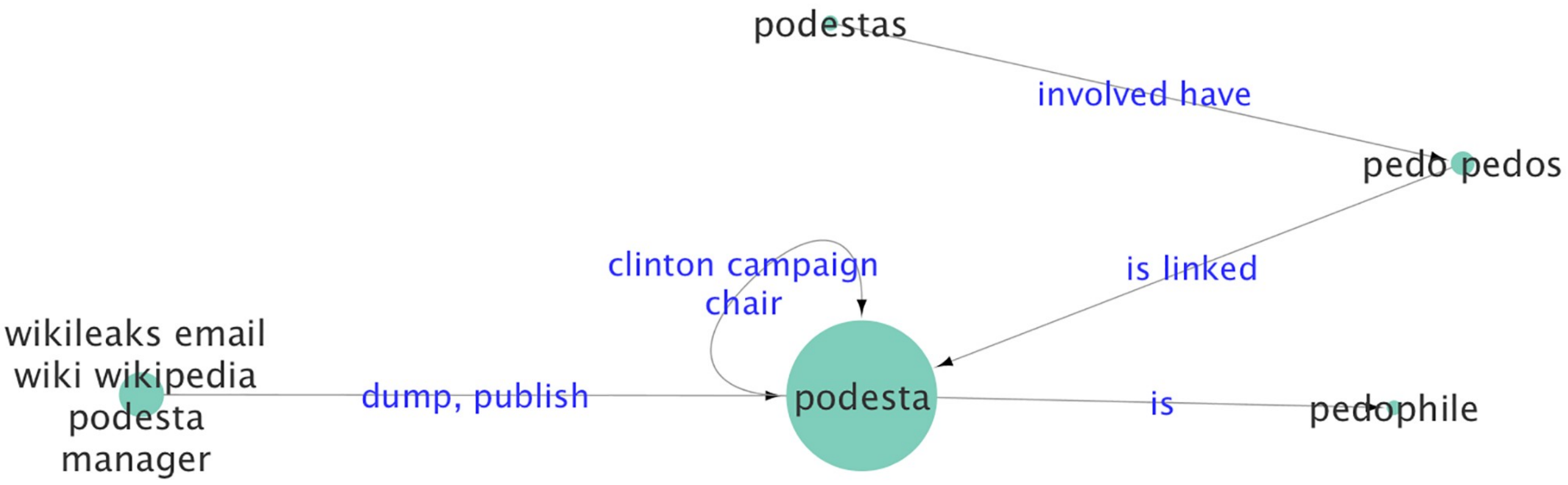

**Fig 12. Selection of nodes from the Podesta subnode egonet subgraph.** The self-loop edge for the node "Podesta" is labeled with an automatically derived description of John Podesta as the Clinton Campaign Chair.

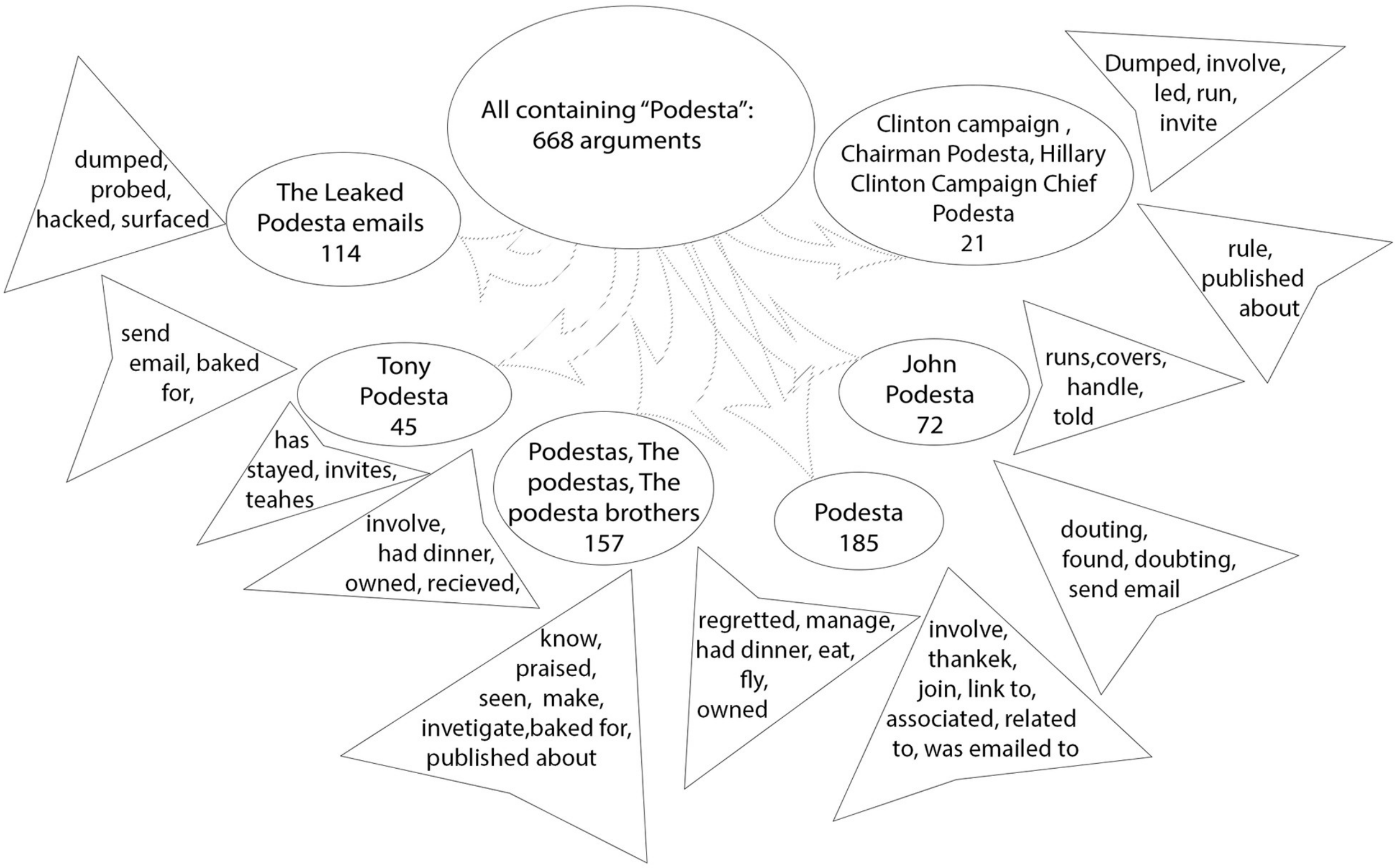

**Fig 13. Subgraph of the Podesta supernode.** The supernode consists of several subnodes, including those automatically labeled as *leaked emails*, *Tony Podesta*, *John Podesta*, *the Podesta brothers*, and *John Podesta as Hillary Clinton's campaign manager*. The most significant context dependent relationships for each of the subnodes are presented as labeled, directed edges. See Fig 14 for further examples where both ends of the relationships are shown.

The automated pipeline for narrative framework discovery provides a clear pathway to developing a sophisticated, multi-scale representation of narratives. Not only does the pipeline capture the top level nodes and relationships such as the ones proposed by the NY Times in their hand-drawn illustrations, but it also captures additional nodes and relationships. For example, our extractions for Pizzagate include important actants such as Bill Clinton and contributions to the Clinton campaign and foundation, which are missing in the NY Times graph but were clearly central to the discussions among Pizzagate conspiracy theorists. Our approach also reveals certain details not captured by the hand-drawn illustrations, such as the central role played by Wikileaks in the Pizzagate conspiracy theory forums in gluing the otherwise disconnected domains of the narrative framework together. Indeed, these findings support our hypothesis that conspiracy theories are built by aligning otherwise unrelated domains of human interaction through the interpretation by the conspiracy theorists of discovered or hidden knowledge to which they claim either to have special access or a particularly astute interpretive ability.

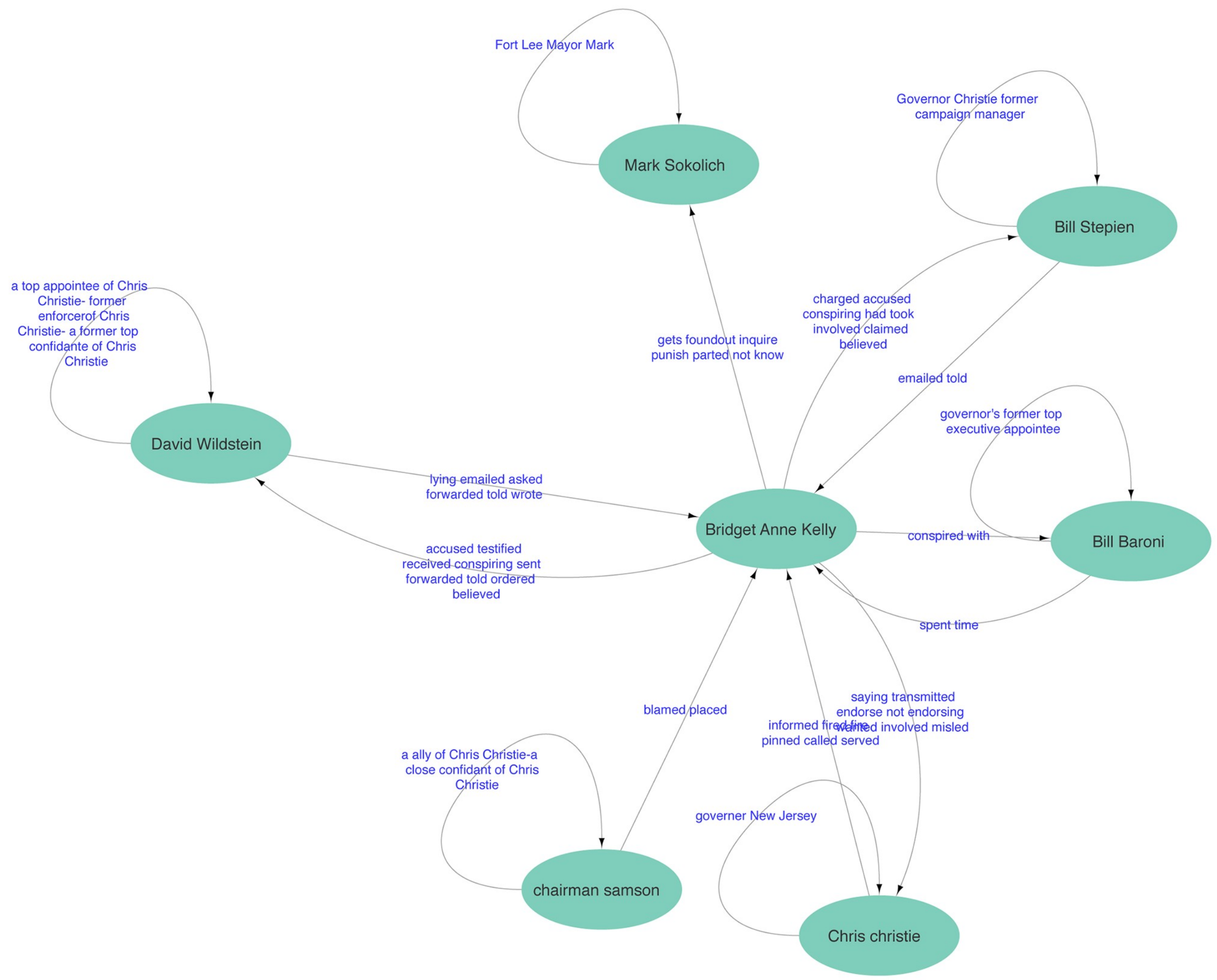

**Fig 14. A subset of the ego network for Bridget Anne Kelly.** The specific relationships between Kelly and other important named individuals are revealed in this ego network subgraph as determined both by their frequency and centrality in the narrative network. For each named entity, we added a self-loop edge labeled with automatically derived descriptions of the entity. These relationships show the endemic nature of the Bridgegate conspiracy: all actants are local to New Jersey politics. Since the edges are labeled with only the lead verbs appearing in the relationship phrases, the edge labels can be difficult to understand but, because we retain the original phrases, the relationship can be recovered. For example, the relationship "pinned" from Christie to Kelly can be linked to sentences such as: *Critchley said evidence to support that claim is contained in interview summaries that accompanied a report commissioned by Christie's office that* ***pinned the blame*** *largely on Kelly and Wildstein.*

An important aspect of our narrative framework discovery is its generative nature. Once the narrative framework is established, one can generate admissible stories or story parts (e.g. forum posts) that conform to the overarching framework by selecting already established actants and relationships. Although such a capacity might be used to create and perpetuate conspiracy theories, it might just as easily be deployed to interrupt narrative frameworks fueling anti-democratic behaviors or encouraging people to take destructive, real-world action. At the very least, our approach allows for deep and powerful insight into story generation, and the underlying factors that allow people to participate in the creation and circulation of these narratives. Similarly, understanding the significant structural differences in narrative frameworks between folkloric genres such as rumors, legends and conspiracy theories on the one hand, and factually reported conspiracies on the other hand, could be useful for testing the veracity of emerging narratives and might prove to be an important component of tools for private and public sector analysts.

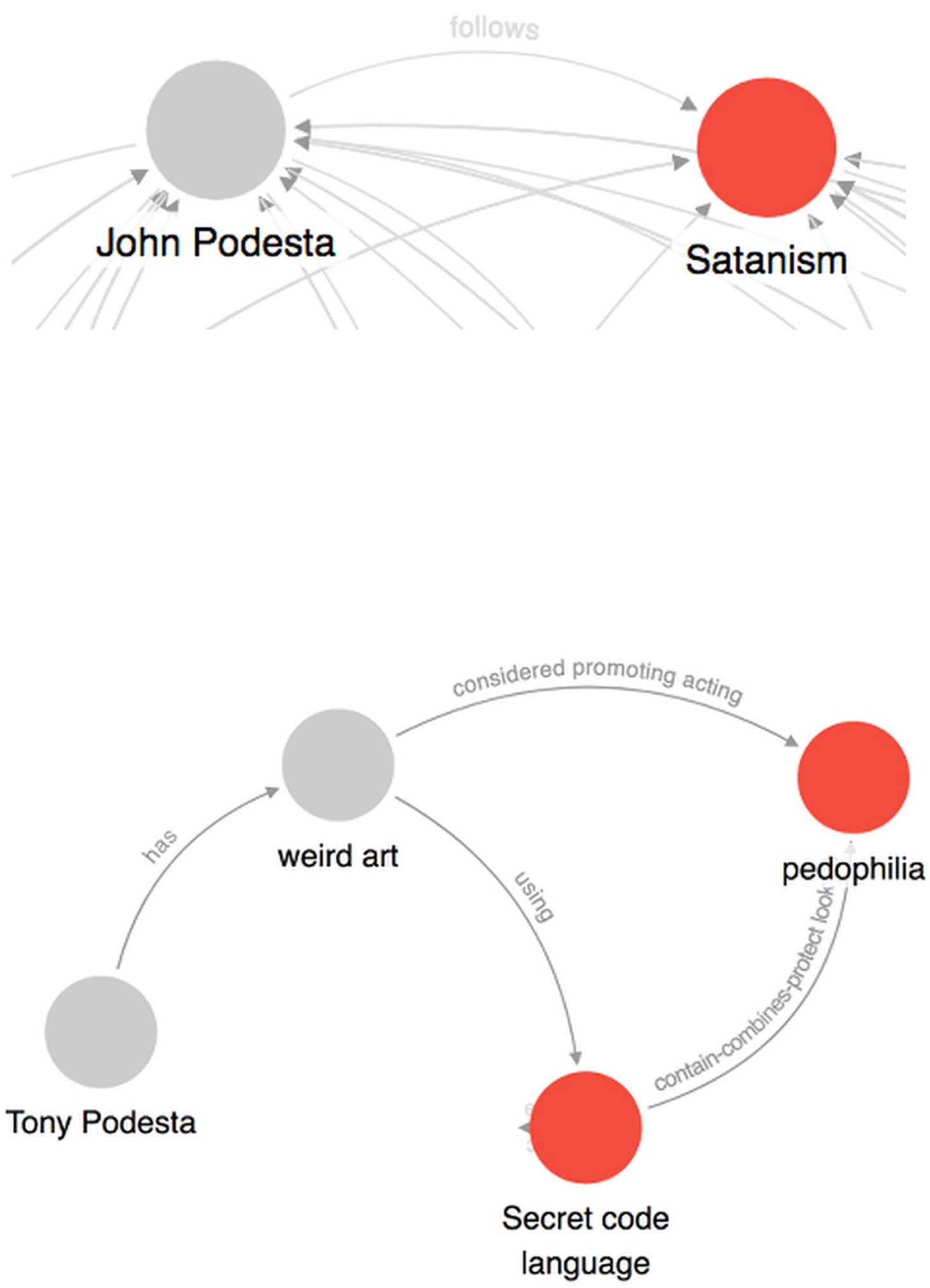

**Fig 15. Two closeups of labeled edges related to Pizzagate.** Excerpts from our auto-generated NY Times matched Pizzagate graph reveal the relationships between a subset of nodes. Top A: the graph reveals that John Podesta follows Satanism, and bottom B: that Tony Podesta owns weird art that uses coded language to promote pedophilia.

## Conclusion

The years of the Trump presidency including the 2016 presidential election, have been marred by what increasingly has come to be known as fake news. Lazer et al propose that fake news be understood as "fabricated information that mimics news media content in form but not in organizational process or intent" [102]. Discerning fact from fiction is difficult given the speed and intensity with which both factual and fictional accounts can spread through both recognized news channels and far more informal social media channels. Accordingly, there is a pressing need, particularly in light of events such as the COVID-19 pandemic, for methods to understand not only how stories circulate on and across these media, but also the generative narrative frameworks on which these stories rest. Recognizing that a series of stories or story fragments align with a narrative framework that has the hallmarks of a fictional conspiracy theory might help counteract the degree to which people come to believe in—and subsequently act on—conspiracy theories.

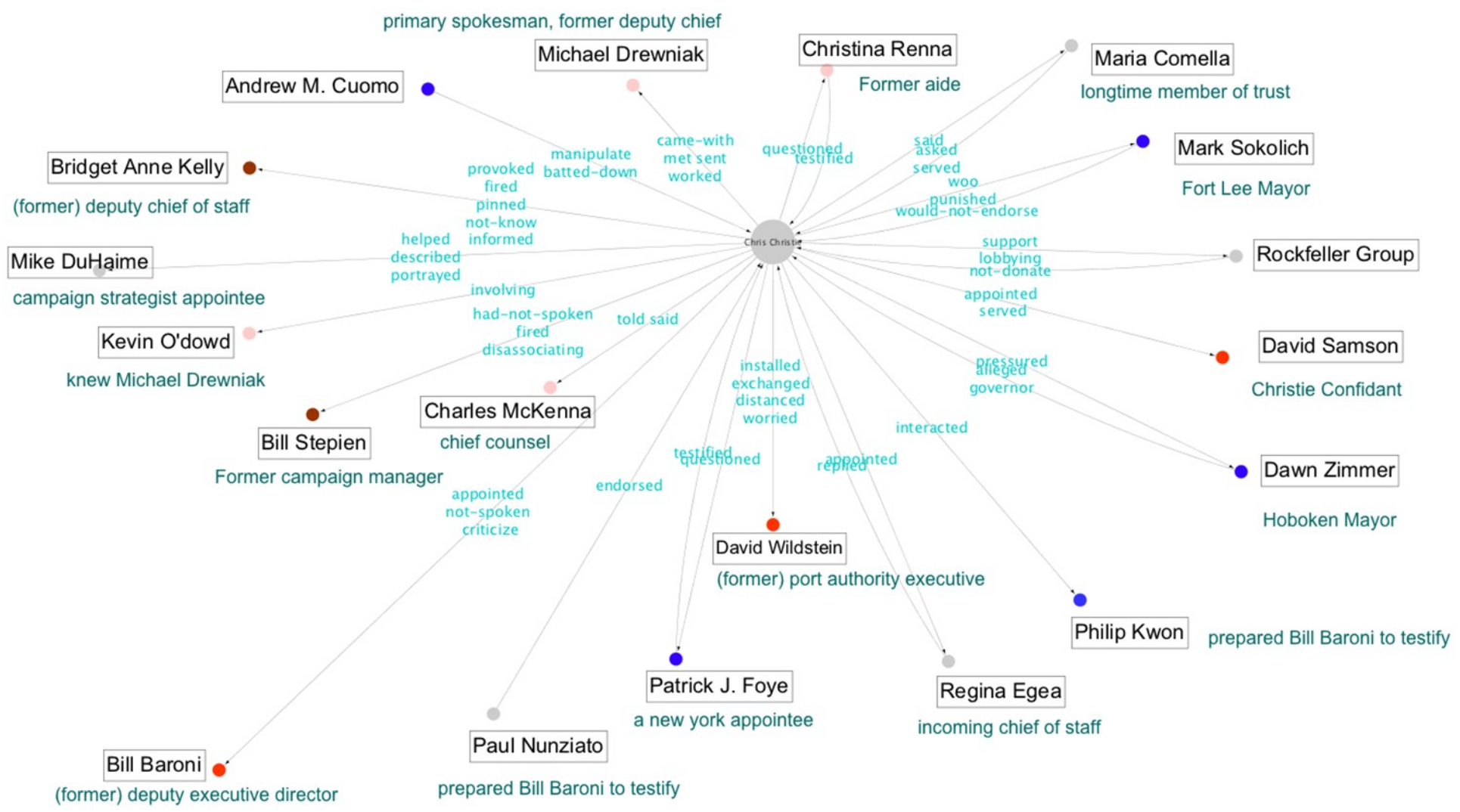

**Fig 16. Comparison of relationship labels generated by our automated methodology with the the NY Times Bridgegate graph for Chris Christie.** Most significant relationship labels from the "Chris Christie" node to other nodes are displayed here. For each node, we also include one descriptive phrase that was found in an automated manner by our pipeline. These descriptive phrases match very closely the roles portrayed in the NY Times Bridgegate graph. As in other figures, the edge labels only pick the most important verbs for the associated relationship phrase. The rest of the words in the corresponding phrases provide the necessary context for meaningful interpretations of these verbs. For example, the verb"pinned" connecting Christie to Anne Bridgett Kelly, is part of the phrase, "pinned the blame on," which we extracted from the text.

We hypothesize that three features—a single domain of interaction, a robustness to deletions of nodes and relationships, and a proliferation of peripheral actants and relationships—are key characteristics of an actual conspiracy and may be helpful in distinguishing actual conspiracies from conspiracy theories. Reporting on actual conspiracies introduces new actants and relationships as part of the process of validating what has actually happened. This reporting feeds the core giant network with more evidence, resulting in a denser network over time. Conspiracy theories, by way of contrast, may form rapidly. Since the only evidence to support any of the actants and relationships comes from the storytellers themselves, we suggest that the network structure of a conspiracy theory stabilizes quickly. This stabilization is supported by studies in folklore, which reveal that an essentially constant and relatively small set of actants and relationships determines the boundaries of admissible stories (or story fragments) after the initial narrative burst finishes [22][57][103]. The addition of new domains through the process of alignment described above and symptomatic of the monological beliefs identified as a common feature of conspiracy theories may at times alter an otherwise stable framework, with sudden changes in the number of actants and relationships included in the network. In short, it seems likely that **a conspiracy theory is characterized by a comparatively small number of actants, multiple interconnected domains, and the fragility of the narrative framework graph, which can easily be disconnected into a series of disjoint subgraphs by the deletion of a small number of nodes or relationships.** Our methods can help derive the narrative frameworks undergirding a corpus, and support the macroscopic analysis of these complex narratives.

Conspiracy theories have in the past been disregarded as the implausible fantasies of fringe members of society, not worthy of serious concern. An increasing awareness that people are making real-world, and at times violent or dangerous, decisions based on informal stories that circulate on and across their social networks, and that conspiracy theories are a significant part of that storytelling, countermands that idea. The rapid spread of conspiracy theories such as Pizzagate, COVID-19 conspiracies, and the capacious QAnon, coupled to the dangerous real world actions that people have taken based on a belief in these narratives, are no longer purely a fringe phenomenon. Consequently, knowledge derived from our methods can have clear and significant public safety impacts, as well as impacts on protecting democratic institutions.

Actual conspiracies and conspiracy theories threaten Democracy each in their own particular way. An actual conspiracy usually comes to light because of the investigative capacities of a free and independent press, and reveals corruption in government or industry; as such, the discovery of an actual conspiracy confirms the power of democratic institutions. Conspiracy theories, on the other hand, seek to undermine the very premise of democratic institutions. As Muirhead and Rosenblum note, "There is no punctilious demand for proofs, no exhausting amassing of evidence, no dots revealed to form a pattern, no close examination of the operators plotting in the shadows. The new conspiracism dispenses with the burden of explanation" [104]. Given the challenges that conspiracy theories present to democracy and a free and open society, we believe that the ability to automatically discover the underlying narrative frameworks for these accounts is of paramount importance. Such an awareness will, at the very least, provide insight into the type of muddled thinking promoted by propaganda campaigns [105] or other disinformation initiatives. It will also offer a clear overview of the domains of knowledge that conspiracy theorists link together through their imaginative interpretations of "hidden knowledge". Identification of the structural aspects of a conspiracy theory narrative framework fueling online conversations, such as the weak connection of multiple domains, can alert us to whether an emerging narrative has the hallmarks of a conspiracy theory. Importantly, these methods can provide insight into the potential strategies that adherents may be considering for dealing with the various threats identified in the narratives. Taken as a whole, the automated narrative framework discovery pipeline can provide us with a better understanding of how stories help influence decision making, and shape the contours of our shifting political environment.

## Supporting information

**S1 File. We provide detailed description of the processing pipeline as presented in Fig 3.** We also provide details of the results that were obtained for the two data sets studied in this paper.
(PDF)

## Acknowledgments

Initial ideas for this project were developed as part of the Institute for Pure and Applied Mathematics long program on Culture Analytics. Earlier versions of this work were presented at Northeastern University's Network Science Institute and NULab for Texts, Maps and Networks, as well as at the University of Vermont's Complex Systems Center. We would like to thank our three annotators for their help in developing our gold standard corpora.

## Author Contributions

**Conceptualization:** Timothy R. Tangherlini, Vwani Roychowdhury.

**Data curation:** Timothy R. Tangherlini, Behnam Shahbazi, Ehsan Ebrahimzadeh.

**Formal analysis:** Shadi Shahsavari, Ehsan Ebrahimzadeh, Vwani Roychowdhury.

**Investigation:** Timothy R. Tangherlini, Shadi Shahsavari, Behnam Shahbazi, Ehsan Ebrahimzadeh, Vwani Roychowdhury.

**Methodology:** Timothy R. Tangherlini, Shadi Shahsavari, Behnam Shahbazi, Ehsan Ebrahimzadeh, Vwani Roychowdhury.

**Resources:** Timothy R. Tangherlini, Vwani Roychowdhury.

**Software:** Timothy R. Tangherlini, Shadi Shahsavari, Behnam Shahbazi, Ehsan Ebrahimzadeh.

**Supervision:** Timothy R. Tangherlini, Vwani Roychowdhury.

**Validation:** Timothy R. Tangherlini, Shadi Shahsavari, Behnam Shahbazi, Vwani Roychowdhury.

**Visualization:** Timothy R. Tangherlini, Behnam Shahbazi, Ehsan Ebrahimzadeh.

**Writing – original draft:** Timothy R. Tangherlini, Shadi Shahsavari, Behnam Shahbazi, Ehsan Ebrahimzadeh, Vwani Roychowdhury.

**Writing – review & editing:** Timothy R. Tangherlini, Vwani Roychowdhury.

## References

1. Rosenblum NL, Muirhead R. A lot of people are saying: the new conspiracism and the assault on democracy. Princeton: Princeton University Press; 2019.
2. Barkun M. Conspiracy theories as stigmatized knowledge. Diogenes. 2016.
3. Oliver JE, Wood TJ. Conspiracy theories and the paranoid style(s) of mass opinion. American Journal of Political Science. 2014; 58(4):952–966. https://doi.org/10.1111/ajps.12084
4. Merlan A. Republic of lies: American conspiracy theorists and their surprising rise to power. New York: Metropolitan Books; 2019.
5. Sunstein CR, Vermeule A. Conspiracy theories: causes and cures. Journal of Political Philosophy. 2009; 17(2):202–227. https://doi.org/10.1111/j.1467-9760.2008.00325.x
6. Goertzel T. Belief in conspiracy theories. Political Psychology. 1994; 15(4):731–742. https://doi.org/10.2307/3791630
7. Hofstadter R. The paranoid style in American politics. New York: Vintage; 2012.
8. Brotherton R, French CC, Pickering AD. Measuring belief in conspiracy theories: the generic conspiracist beliefs scale. Frontiers in psychology. 2013; 4:279.
9. van Prooijen JW, Acker M. The influence of control on belief in conspiracy theories: conceptual and applied extensions. Applied Cognitive Psychology. 2015; 29(5):753–761. https://doi.org/10.1002/acp.3161
10. Swami V, Furnham A, Smyth N, Weis L, Lay A, Clow A. Putting the stress on conspiracy theories: examining associations between psychological stress, anxiety, and belief in conspiracy theories. Personality and Individual Differences. 2016; 99:72–76. https://doi.org/10.1016/j.paid.2016.04.084
11. Douglas KM, Sutton RM, Cichocka A. The psychology of conspiracy theories. Current directions in psychological science. 2017; 26(6):538–542. https://doi.org/10.1177/0963721417718261
12. Pennycook G, Cannon TD, Rand DG. Prior exposure increases perceived accuracy of fake news. Journal of experimental psychology: general. 2018; 147(12):1865. https://doi.org/10.1037/xge0000465
13. Raab MH, Auer N, Ortlieb SA, Carbon CC. The Sarrazin effect: the presence of absurd statements in conspiracy theories makes canonical information less plausible. Frontiers in psychology. 2013; 4:453.
14. Raab MH, Ortlieb S, Auer N, Guthmann K, Carbon CC. Thirty shades of truth: conspiracy theories as stories of individuation, not of pathological delusion. Frontiers in psychology. 2013; 4:406.

15. Swami V, Coles R, Stieger S, Pietschnig J, Furnham A, Rehim S, et al. Conspiracist ideation in Britain and Austria: evidence of a monological belief system and associations between individual psychological differences and real-world and fictitious conspiracy theories. British Journal of Psychology. 2011; 102(3):443–463. https://doi.org/10.1111/j.2044-8295.2010.02004.x PMID: 21751999

16. Fenster M. Conspiracy theories: secrecy and power in American culture. Minneapolis: University of Minnesota Press; 1999.

17. Rosnow RL, Fine GA. Rumor and gossip: the social psychology of hearsay. New York: Elsevier; 1976.

18. Grimes DR. On the viability of conspiratorial beliefs. PloS one. 2016; 11(1):e0147905. https://doi.org/10.1371/journal.pone.0147905

19. Samory M, Mitra T. Conspiracies online: user discussions in a conspiracy community following dramatic events. In: Twelfth International AAAI Conference on Web and Social Media; 2018.

20. Klein C, Clutton P, Dunn AG. Pathways to conspiracy: the social and linguistic precursors of involvement in Reddit's conspiracy theory forum. PloS one. 2019; 14(11):e0225098. https://doi.org/10.1371/journal.pone.0225098

21. Ellis B. Raising the devil: Satanism, new religious movements, and the media. Louisville: University Press of Kentucky; 2000.

22. Campion-Vincent V, Renard JB. Conspiracy theories today. Diogenes. 2015; 249-250(1):1–264.

23. Knight P. Conspiracy theories in American history: an encyclopedia. vol. 1. Santa Barbara: ABC-CLIO; 2003.

24. Arnold GB. Conspiracy theory in film, television, and politics. Santa Barbara: ABC-CLIO; 2008.

25. Graumann CF, Moscovici S. Changing conceptions of conspiracy. New York: Springer-Verlag; 1987.

26. Metaxas P, Finn ST. The infamous#Pizzagate conspiracy theory: insight from a TwitterTrails investigation. Wellesley College Faculty Research and Scholarship. 2017; 188. Available from: https://repository.wellesley.edu/scholarship/188

27. Ferrara E, Varol O, Davis C, Menczer F, Flammini A. The rise of social bots. Communications of the ACM. 2016; 59(7):96–104. https://doi.org/10.1145/2818717

28. Del Vicario M, Bessi A, Zollo F, Petroni F, Scala A, Caldarelli G, et al. The spreading of misinformation online. Proceedings of the National Academy of Sciences. 2016; 113(3):554–559. https://doi.org/10.1073/pnas.1517441113

29. Samory M, Mitra T. 'The government spies using our webcams': the language of conspiracy theories in online discussions. Proceedings of the ACM on Human-Computer Interaction. 2018; 2(CSCW):1–24.

30. Maddock J, Starbird K, Al-Hassani HJ, Sandoval DE, Orand M, Mason RM. Characterizing online rumoring behavior using multi-dimensional signatures. In: Proceedings of the 18th ACM conference on computer supported cooperative work & social computing; 2015. pp. 228–241.

31. Bessi A, Coletto M, Davidescu GA, Scala A, Caldarelli G, Quattrociocchi W. Science vs conspiracy: collective narratives in the age of misinformation. PloS one. 2015; 10(2):e0118093. https://doi.org/10.1371/journal.pone.0118093

32. Van der Linden S. The conspiracy-effect: exposure to conspiracy theories (about global warming) decreases pro-social behavior and science acceptance. Personality and Individual Differences. 2015; 87:171–173. https://doi.org/10.1016/j.paid.2015.07.045

33. Jolley D, Douglas KM. The effects of anti-vaccine conspiracy theories on vaccination intentions. PloS one. 2014; 9(2):e89177. https://doi.org/10.1371/journal.pone.0089177

34. Barkun M. A culture of conspiracy: apocalyptic visions in contemporary America. vol. 15. Berkeley: University of California Press; 2013.

35. Tangherlini TR, Roychowdhury V, Glenn B, Crespi CM, Bandari R, Wadia A, et al. "Mommy blogs" and the vaccination exemption narrative: results from a machine-learning approach for story aggregation on parenting social media sites. JMIR public health and surveillance. 2016; 2(2):e166. https://doi.org/10.2196/publichealth.6586 PMID: 27876690

36. Ohlheiser A. Fearing yet another witch hunt, Reddit bans 'Pizzagate'. The Washington Post. 2016 Nov 24 [Cited 2020 May 2]. Available from: https://www.washingtonpost.com/news/the-intersect/wp/2016/11/23/fearing-yet-another-witch-hunt-reddit-bans-pizzagate/

37. Torres-Soriano MR. The dynamics of the creation, evolution, and disappearance of terrorist Internet forums. International Journal of Conflict and Violence (IJCV). 2013; 7(1):164–178.

38. Villegas EB. Facebook and its disappearing posts: data collection approaches on fan-pages for social scientists. The Journal of Social Media in Society. 2016; 5(1):160–188.

39. Tangherlini T. Toward a generative model of legend: pizzas, bridges, vaccines, and witches. Humanities. 2018; 7(1):1.
40. Dégh L. What is a belief legend? Folklore. 1996; 107(1-2):33–46.
41. Dvorak P. At a D.C. pizzeria, the dangers of fake news just got all too real. Washington Post. 2016 Dec 5 [Cited 2020 May 2]. Available from: https://www.washingtonpost.com/local/at-a-dc-pizzeria-the-dangers-of-fake-news-just-got-all-too-real/2016/12/05/b8ae43b8-baf4-11e6-94ac-3d324840106c_story.html
42. Bearman P, Stovel K. Becoming a Nazi: a model for narrative networks. Poetics 2000; 27(2-3):69–90. https://doi.org/10.1016/S0304-422X(99)00022-4
43. Lehnert WG. Narrative text summarization. In: AAAI 1980. pp. 337–339.
44. Boole G. An investigation of the laws of thought: on which are founded the mathematical theories of logic and probabilities. London: Walton and Maberly; 1854.
45. Fine GA, Campion-Vincent V, Heath C. Rumor mills: the social impact of rumor and legend. New Brunswick: Aldine Transaction; 2005.
46. Bandari R, Zhou Z, Qian H, Tangherlini TR, Roychowdhury VP. A resistant strain: revealing the online grassroots rise of the antivaccination movement. Computer. 2017; 50(11):60–67. https://doi.org/10.1109/MC.2017.4041354
47. Greimas AJ. Éléments pour une théorie de l'interprétation du récit mythique. Communications. 1966; 8(1):28–59.
48. Greimas A. Sémantique structurale. Paris: Larousse; 1966.
49. Greimas AJ. Les actants, les acteurs et les figures. Sémiotique narrative et textuelle. 1973; pp. 161–176.
50. Todorov T. Les catégories du récit littéraire. Communications. 1966; 8(1):125–151.
51. Eskeröd A. Årets äring: etnologiska studier i skördens och julens tro och sed. Nordiska Museets Handlingar 6. Stockholm: Nordiska Museet; 1947.
52. Börner K. Plug-and-play macroscopes. Communications of the ACM. 2011; 54(3):60–9. https://doi.org/10.1145/1897852.1897871
53. Duncan CA, Eppstein D, Goodrich MT, Kobourov SG, Nöllenburg M. Lombardi drawings of graphs. J Graph Algorithms Appl. 2012; 16(1):85–108. https://doi.org/10.7155/jgaa.00251
54. Kennicott P. Art can help distinguish between conspiracy and reality, and this exhibition proves it. The Washington Post. 2018 Nov 4 [Cited 2020 Mar 3]. Available from: https://www.washingtonpost.com/entertainment/museums/art-can-help-distinguish-between-conspiracy-and-reality-and-this-exhibition-proves-it/2018/11/03/e548a8f0-ded3-11e8-b3f0-62607289efee_story.html
55. Aisch G, Huang J, Kang C. Dissecting the PizzaGate conspiracy theories. New York Times. 2016 Dec 10 [Cited 2020 Mar 3]. Available from: https://www.nytimes.com/interactive/2016/12/10/business/media/pizzagate.html
56. Marsh B, Zernike K. Chris Christie and the lane closings: a spectator's guide. The New York Times. 2015 Apr 8 [Cited 2020 Mar 3]. Available from: https://www.nytimes.com/interactive/2015/04/08/nyregion/chris-christie-and-bridgegate-guide.html
57. Anderson W. Kaiser und Abt: die Geschichte eines Schwanks. FF communication 42. Helsinki: Suomalainen Tiedeakatemia; 1923.
58. Clover CJ. The long prose form. Arkiv för nordisk filologi. 1986; 101:10–39.
59. Laudun J. Talk about the past in a midwestern town:"It was there at that time.". Midwestern Folklore. 2001; 27(2):41–54.
60. Goertzel B. Chaotic logic: language, thought, and reality from the perspective of complex systems science. New York: Plenum Press; 1994.
61. Falahi M. A cognition-driven approach to modeling document generation and learning underlying contexts from documents. Ph.D. Dissertation, UCLA. 2017. Available from: https://escholarship.org/uc/item/8ft505bg
62. Morzy M. On mining and social role discovery in Internet forums. In: 2009 International Workshop on Social Informatics. IEEE; 2009. pp. 74–79.
63. Klein C, Clutton P, Polito V. Topic modeling reveals distinct interests within an online conspiracy forum. Frontiers in psychology. 2018; 9:189. https://doi.org/10.3389/fpsyg.2018.00189
64. Gildea D, Jurafsky D. Automatic labeling of semantic roles. Computational linguistics. 2002; 28(3):245–288. https://doi.org/10.1162/089120102760275983

65. Manning C, Surdeanu M, Bauer J, Finkel J, Bethard S, McClosky D. The Stanford CoreNLP natural language processing toolkit. In: Proceedings of 52nd annual meeting of the association for computational linguistics: system demonstrations; 2014. pp. 55–60.

66. Akbik A, Bergmann T, Blythe D, Rasul K, Schweter S, Vollgraf R. FLAIR: an easy-to-use framework for state-of-the-art nlp. In: Proceedings of the 2019 Conference of the North American Chapter of the Association for Computational Linguistics (Demonstrations); 2019. pp. 54–59.

67. Pennington J, Socher R, Manning C. Glove: global vectors for word representation. In: Proceedings of the 2014 conference on empirical methods in natural language processing (EMNLP); 2014. pp. 1532–1543.

68. Mikolov T, Sutskever I, Chen K, Corrado GS, Dean J. Distributed representations of words and phrases and their compositionality. In: Advances in neural information processing systems; 2013. pp. 3111–3119.

69. Bird S, Loper E. NLTK: the natural language toolkit. In: Proceedings of the ACL 2004 on Interactive poster and demonstration sessions. Association for Computational Linguistics; 2004. p. 31.

70. Collobert R, Weston J, Bottou L, Karlen M, Kavukcuoglu K, Kuksa P. Natural language processing (almost) from scratch. Journal of Machine Learning Research. 2011; 12(Aug):2493–2537.

71. De Marneffe MC, Dozat T, Silveira N, Haverinen K, Ginter F, Nivre J, et al. Universal Stanford dependencies: a cross-linguistic typology. In: LREC. vol. 14; 2014. pp. 4585–4592.

72. de Marneffe MC, Connor M, Silveira N, Bowman SR, Dozat T, Manning CD. More constructions, more genres: extending Stanford dependencies. In: Proceedings of the Second International Conference on Dependency Linguistics (DepLing 2013); 2013. pp. 187–196.

73. Schuster S, Manning CD. Enhanced English universal dependencies: an improved representation for natural language understanding tasks. In: LREC; 2016. pp. 2371-2378.

74. Baker CF, Fillmore CJ, Lowe JB. The Berkeley framenet project. In: Proceedings of the 17th international conference on Computational linguistics-Volume 1. Association for Computational Linguistics; 1998. pp. 86–90.

75. Palmer M, Gildea D, Kingsbury P. The proposition bank: an annotated corpus of semantic roles. Computational linguistics. 2005; 31(1):71–106. https://doi.org/10.1162/0891201053630264

76. Schmitz M, Bart R, Soderland S, Etzioni O. Open language learning for information extraction. In: Proceedings of the 2012 Joint Conference on Empirical Methods in Natural Language Processing and Computational Natural Language Learning. Association for Computational Linguistics; 2012. pp. 523–534.

77. Del Corro L, Gemulla R. Clausie: clause-based open information extraction. In: Proceedings of the 22nd international conference on World Wide Web. ACM; 2013. pp. 355–366.

78. Fader A, Soderland S, Etzioni O. Identifying relations for open information extraction. In: Proceedings of the conference on empirical methods in natural language processing. Association for Computational Linguistics; 2011. pp. 1535–1545.

79. Wu F, Weld DS. Open information extraction using Wikipedia. In: Proceedings of the 48th Annual Meeting of the Association for Computational Linguistics. Association for Computational Linguistics; 2010. pp. 118–127.

80. Mohr JW, Wagner-Pacifici R, Breiger RL, Bogdanov P. Graphing the grammar of motives in national security strategies: cultural interpretation, automated text analysis and the drama of global politics. Poetics. 2013; 41(6):670–700. https://doi.org/10.1016/j.poetic.2013.08.003

81. Devlin J, Chang MW, Lee K, Toutanova K. Bert: Pre-training of deep bidirectional transformers for language understanding. arXiv:181004805 [Preprint]. 2018 [cited 2020 March 3]. Available from: https://arxiv.org/abs/1810.04805

82. Arthur D, Vassilvitskii S. k-means++: the advantages of careful seeding. In: Proceedings of the eighteenth annual ACM-SIAM symposium on Discrete algorithms. Society for Industrial and Applied Mathematics; 2007. pp. 1027–1035.

83. Bellegarda JR, Butzberger JW, Chow YL, Coccaro NB, Naik D. A novel word clustering algorithm based on latent semantic analysis. In: 1996 IEEE International Conference on Acoustics, Speech, and Signal Processing Conference Proceedings. vol. 1. IEEE; 1996. pp. 172–175.

84. Bigi B. Using Kullback-Leibler distance for text categorization. In: European Conference on Information Retrieval. Berlin: Springer; 2003. pp. 305–319.

85. Newman ME. Clustering and preferential attachment in growing networks. Physical review E. 2001; 64 (2):025102. https://doi.org/10.1103/PhysRevE.64.025102

86. Fortunato S. Community detection in graphs. Physics reports. 2010; 486(3-5): 75–174. https://doi.org/10.1016/j.physrep.2009.11.002

87. Newman ME, Girvan M. Finding and evaluating community structure in networks. Physical review E. 2004; 69(2):026114.

88. Aynaud T. Community detection for NetworkX's documentation. 2018. Available from: https://networkx.github.io/documentation/stable/reference/algorithms/community.html

89. Hagberg A, Swart P, Chult DS. Exploring network structure, dynamics, and function using NetworkX. Los Alamos National Lab.(LANL), Los Alamos, NM (United States); 2008.

90. De Domenico M, Porter MA, Arenas A. MuxViz: a tool for multilayer analysis and visualization of networks. Journal of Complex Networks. 2015; 3(2):159–176. https://doi.org/10.1093/comnet/cnu038

91. Skomorovsky M, Mintz M. Oligrapher. 2019 [Cited 2020 Mar 3]. Available from: https://github.com/public-accountability/oligrapher.

92. Shannon P, Markiel A, Ozier O, Baliga NS, Wang JT, Ramage D, Amin N, Schwikowski B, Ideker T. Cytoscape: a software environment for integrated models of biomolecular interaction networks. Genome research. 2003 Nov 1; 13(11):2498–504. https://doi.org/10.1101/gr.1239303

93. Bastian M, Heymann S, Jacomy M. Gephi: an open source software for exploring and manipulating networks. 2009. Available from: http://www.aaai.org/ocs/index.php/ICWSM/09/paper/view/154.

94. Zafarani R, Liu H. Evaluation without ground truth in social media research. Communications of the ACM. 2015; 58(6):54–60. https://doi.org/10.1145/2666680

95. Waisbord S. Truth is what happens to news: on journalism, fake news, and post-truth. Journalism studies. 2018; 19(13):1866–78. https://doi.org/10.1080/1461670X.2018.1492881

96. Wissler L, Almashraee M, Díaz DM, Paschke A. The gold standard in corpus annotation. In: IEEE GSC 2014.

97. Gius E, Reiter N, Willand M. Foreword to the special issue "A shared task for the digital humanities: annotating narrative levels". Journal of Cultural Analytics. 2019.

98. Artstein R. Inter-annotator agreement. In: Ide N, Pustejovsky J, editors. Handbook of linguistic annotation. Berlin: Springer; 2017. pp. 297–313.

99. Uzuner Ö, Solti I, Xia F, Cadag E. Community annotation experiment for ground truth generation for the i2b2 medication challenge. Journal of the American Medical Informatics Association. 2010; 17(5):519–23. https://doi.org/10.1136/jamia.2010.004200 PMID: 20819855

100. Zuckerman E. QAnon and the emergence of the unreal. Journal of Design and Science. 2019; 15(6).

101. Abello J, Van Ham F, Krishnan N. Ask-graphview: a large scale graph visualization system. IEEE transactions on visualization and computer graphics. 2006; 12(5):669–76. https://doi.org/10.1109/TVCG.2006.120

102. Lazer DM, Baum MA, Benkler Y, Berinsky AJ, Greenhill KM, Menczer F, et al. The science of fake news. Science. 2018; 359(6380):1094–1096. https://doi.org/10.1126/science.aao2998 PMID: 29590025

103. Starbird K. Information wars: a window into the alternative media ecosystem. Medium. 2017 Mar 14 [Cited 2020 Mar 3]. Available from: https://medium.com/hci-design-at-uw/information-wars-a-window-into-the-alternative-media-ecosystem-a1347f32fd8f

104. Muirhead R, Rosenblum N. The New Conspiracists. Dissent. 2018; 65(1):51–60. https://doi.org/10.1353/dss.2018.0012

105. Pomerantsev P, Weiss M. The menace of unreality: how the Kremlin weaponizes information, culture and money. New York: Institute of Modern Russia; 2014.